# Automating Geometry-Intensive Compliance Checking in BIM: Graph-Based Semantic Reasoning Framework

Zixuan Xiao[1], Pei Troh Koh[1], Jun Ma[1,*], Jack C.P. Cheng[2]

[1] Department of Urban Planning and Design, The University of Hong Kong, Hong Kong
[2] Department of Civil and Environmental Engineering, The Hong Kong University of Science and Technology, Hong Kong
[*] Corresponding author (Email address: junma@hku.hk)

**Abstract**

Automating compliance check for geometry-intensive regulations remains a significant technical bottleneck in Building Information Modeling (BIM), primarily due to the semantic disparity between high-level regulatory logic and structured IFC data. Existing methods, often reliant on static rule templates, struggle to traverse multi-hop reasoning chains or resolve latent spatial dependencies across multiple building entities. To address these challenges, a Spatial-Geometric Reasoning System for Building Information Modeling (SGR-BIM) is proposed as an integrative graph-driven reasoning framework. SGR-BIM dynamically constructs a cross-modal knowledge graph that aligns user intent, regulatory semantics, and BIM geometry, enabling interpretable reasoning without rigid hard-coding. Validated on 679 expert-verified queries from fire safety codes, the framework achieves 84.3% accuracy, representing an 8.6% improvement over enhanced-tool single-agent baselines. This research provides a graph-based semantic reasoning paradigm, enhancing the transparency and flexibility of automated geometric compliance check workflows in the Architecture, Engineering, and Construction (AEC) industry.

**Keywords:** Automated Compliance Check; Building Information Modeling (BIM); Spatial Reasoning; Large Language Model (LLM);

## 1. Introduction

Automated Compliance Check (ACC) has become a pivotal research topic in the Architecture, Engineering, and Construction (AEC) industry, aiming to streamline regulatory enforcement and reduce manual errors in design review. However, a persistent challenge in current ACC workflows is the automated verification of geometry-intensive regulations, rules that rely on complex spatial relationships rather than simple attribute matching. In this study, geometry-intensive regulations specifically refer to regulatory provisions that are evaluated through semantic geometric properties (e.g., parametric dimensions) and topological relationships explicitly or implicitly defined within the IFC schema. While IFC supports various representations such as B-rep or CSG, this research focuses on semantic-level geometric reasoning rather than mesh-to-mesh Boolean operations.

Fire safety codes represent a quintessential example of such computational complexity. While critical for safeguarding public safety [1,2], these codes are inherently geometry-dependent, prescribing strict requirements on compartmentation, means of escape, and evacuation capacity. Unlike simple property checks, verifying these provisions requires interpreting intricate spatial parameters such as compartment area, travel distance, stair and corridor widths, clearances, and door opening dimensions. Ensuring compliance with these geometry-intensive rules is therefore not only a safety imperative but also a significant technical benchmark for advancing automated design review.

The increasing adoption of Building Information Modeling (BIM) provides new opportunities for automating the compliance checking of fire safety regulations. IFC-based BIM models capture rich geometric and semantic information that can, in principle, support the automatic evaluation of code provisions. Prior studies have explored various methods including rule-based systems, logic and constraint checking, model querying, and ontology-driven reasoning to translate regulatory requirements into machine-executable representations [3,4]. These approaches have shown promising potential for checking relatively simple attribute-level or condition-level rules, such as material types or element properties [5,6].

However, geometry compliance check remains an unresolved and particularly challenging problem, despite the increasing availability of IFC-based BIM models and growing interest in automating regulatory evaluation. A major difficulty arises from the multi-hop and context-dependent nature of geometric rule chains embedded in regulatory codes. Many geometry-related provisions cannot be validated through a single geometric attribute. Instead, they require sequential interpretation of multiple interlinked clauses across different sections of the regulation. For instance, verifying the number and width of exit doors represents a typical multi-hop rule chain challenge. This task is not a standalone geometric check, which requires a sequential reasoning path by determining the building’s Use Classification to identify the correct Occupancy Factor, which is then combined with room sizes to compute the Occupant Capacity.

Existing rule-based or constraint-check approaches have advanced the field through semi-automated rule translation using visual programming [7], comparative frameworks for Linked Data and IFC-based validation [8], and the use of Large Language Models (LLMs) for automated regulatory text interpretation [9,10]. However, these approaches, which primarily rely on static rule templates or hard-coded logic, are not well-suited for traversing such dynamic, interlinked logical dependencies across different regulatory sections. Specifically, existing template-based systems fail in such scenarios because they typically presuppose a static mapping between a rule and a target attribute (e.g., Door.Width > Constant). They lack the computational flexibility to dynamically execute a pre-requisite reasoning chain, such as querying occupancy tables and calculating area sums, before performing the geometric check. Consequently, a user would be required to manually pre-calculate and input the 'Occupant Capacity' into the BIM model, which defeats the purpose of full automation and introduces significant potential for human error.

In addition to the inherent multi-step nature of geometric rules, a more fundamental challenge arises from the semantic gap among user queries, code provisions, and BIM geometry. In this study, we define the semantic gap as the disparity between the high-level linguistic logic of regulations and the low-level data structures of IFC. Consequently, cross-modal alignment is the process of mapping these heterogeneous sources, user intent, regulatory text, and BIM data, into a unified logical reasoning space. These three information sources operate at different levels of abstraction and are rarely aligned in practice. This gap is manifested in underspecified user queries. For example, a query like 'Does this stair comply with the code?' requires the system to first infer from the regulatory text which specific geometric attributes (e.g., width) must be validated. Furthermore, many regulatory requirements, which combine narrative natural language with structured and semi-structured tabular specifications that enumerate dimensional thresholds, classification criteria, and exception conditions, involve latent elements linkage. Verifying an exit door’s width often necessitates retrieving information from the IFCBuilding (for Use Classification) and IFCSpace (for room area), rather than the door entity alone. Without a mechanism for precise and transparent cross-modal alignment across these modalities, automated methods cannot reliably support comprehensive and explainable geometry-based fire safety evaluations.

To address these cross-modal alignment challenges, this study proposes SGR-BIM (Spatial-Geometric Reasoning System for BIM), a graph-based semantic reasoning framework designed to automate the verification of geometry-intensive regulations. Unlike traditional pipelines that rely on static rule templates, SGR-BIM introduces a dynamic cross-modal knowledge graph, which unifies diverse information sources, including user intent, regulatory semantics, and IFC-based geometric data, within a coherent topological representation. By embedding spatial relationships directly into the graph structure, the framework enables the system to autonomously interpret multi-hop rule chains, resolve latent dependencies across building entities, and extract context-specific spatial attributes required for high-precision evaluation. On the other hand, SGR-BIM orchestrates this process through coordinated semantic reasoning modules that decompose complex spatial tasks into transparent sub-tasks. Every decision generates an interpretable reasoning trace, allowing practitioners to validate the underlying geometric assumptions.

This study is built upon the hypothesis that the coordination of specialized agents through a dynamic graph structure can effectively bridge the semantic gap by aligning heterogeneous information sources into a unified reasoning path, thereby resolving multi-hop spatial dependencies that remain insurmountable for static rule-template systems. To systematically evaluate this hypothesis through the SGR-BIM framework, we formulate three Research Questions (RQs): (RQ1) How effectively can the framework align user queries, regulatory codes, and BIM models, to provide effective and accurate automated geometry-intensive compliance checks? (RQ2) to what extent can the graph-driven mechanism provide users with complete and relevant answers? and (RQ3) How does the generated reasoning path contribute to the interpretability of the results?

Building upon the above research questions, the objective of this study is to develop and validate a computational framework that can (1) accurately align heterogeneous information sources across user queries, regulatory codes, and BIM models, (2) effectively resolve multi-hop and context-dependent geometric rule chains, and (3) provide interpretable reasoning processes that support transparent compliance checking. These objectives are directly mapped with four evaluation metrics: Accuracy, Coherence, Relevance, and Explainability. Accuracy serves as the primary quantitative measure to validate the fulfilment of RQ1, reflecting the system's factual alignment ability with the three information modalities. The quality of the response in RQ2 is captured by Relevance and Coherence, which together evaluate whether the aligned context is both substantively related to the user's intent and logically complete. Finally, the Explainability metric is mapped to RQ3 to assess the transparency of the logic chain, ensuring the reasoning paths meet the interpretability standards required for professional regulatory workflows.

The main contributions of this paper are summarized as follows:

(1) SGR-BIM, an integrative framework for automated spatial constraint verification, is proposed to overcome the limitations of hard-coded logic in handling context-dependent geometric rules.

(2) A graph-driven cross-modal alignment mechanism is introduced to dynamically bridge the semantic gap between unstructured regulatory texts, user queries, and structured IFC geometry, enabling unified multi-hop reasoning.

(3) An explainable and interactive compliance-checking pipeline is designed, integrating a backend checking system with an interactive user interface to support transparent, query-driven spatial evaluation.

The remainder of this paper is organized as follows: Section 2 reviews related work in automated compliance checking and the application of Large Language Models and agents in the AEC industry. Section 3 delineates the methodology of the proposed SGR-BIM framework, detailing the graph-driven cross-modal structured semantic alignment and the interactive interface. Section 4 documents the experimental setup and provides a comprehensive evaluation of quantitative performance, followed by a diagnostic error analysis and a multifaceted discussion on knowledge graph characteristics and reasoning quality. Section 5 acknowledges the research limitations and identifies specific directions for future work. Finally, Section 6 summarizes the core contributions and concludes the study.

## 2. Related Work

This section situates the proposed SGR-BIM framework within the broader landscape of digital construction and artificial intelligence. Section 2.1 reviews the evolution of Automated Compliance Checking (ACC) systems. Section 2.2 explores the recent advancements in Large Language Models (LLMs) and multi-agent reasoning frameworks, focusing on their emerging applications within the Architecture, Engineering, and Construction (AEC) industry.

### 2.1. Automated Compliance Check for Building Regulations

Automated compliance check has been a prominent research topic in the AEC domain for more than two decades, aiming to reduce the cost, inconsistency, and inefficiency associated with manual building code review. Existing approaches span a methodological spectrum from rule- and logic-based formalisms, to Building Information Modeling (BIM) query and model-driven validation workflows, and more recently to Natural Language Processing (NLP)- or knowledge-based regulation interpretation. Together, these research directions illustrate the evolution of automated compliance check from manually authored, rigid rule representations toward more flexible, data-driven, and semi-intelligent pipelines. This trajectory is exemplified by the integration of knowledge graphs for semantic rule modeling [11] and has been extensively documented in meta-analyses and systematic reviews of the field's methodological shifts [12,13].

Rule- and logic-based approaches represent the earliest and most extensively studied category of automated compliance check methods [14,15]. These systems formalize regulatory requirements into machine-executable expressions through manually crafted rule templates, procedural scripts, or declarative logic rules. To improve semantic expressiveness and reuse, ontology-driven methods encode regulatory knowledge using Web Ontology Language (OWL) / Resource Description Framework (RDF) and evaluate constraints through Semantic Web Rule Language (SWRL) or SPARQL Protocol and RDF Query Language (SPARQL) inference [16,17]. Although such approaches offer high interpretability and explicit control over rule semantics, they require substantial expert effort to author and maintain rule sets, and their expressiveness is constrained by formal logic constructs. Due to the open-world assumption of OWL/SWRL and the limited support for sequencing dependencies, these systems struggle to represent multi-hop and context-dependent rule chains commonly found in fire safety codes.

BIM query- and model-driven approaches shift the focus toward direct validation of IFC-based BIM models [18,19]. These methods leverage model querying, schema-based validation (e.g., JSON/XML Schema,

Information Delivery Specification (IDS)), and geometric or spatial computation to assess compliance. Visual rule authoring languages such as Visual Code Checking Language (VCCL) further lower the barrier for composing rule constraints, while hybrid query–geometry systems have demonstrated automated evaluation of egress paths, fire compartment boundaries, thermal performance, visibility, and accessibility conditions [20,21]. Such approaches align well with BIM-centric workflows and can handle geometric or spatial conditions that exceed the capabilities of pure logic systems. However, their effectiveness depends heavily on data completeness and consistent semantic modeling within BIM. More critically, they are typically limited to single-condition or single-object evaluation and lack mechanisms to handle latent linkage between multiple IFC entities, which is essential for geometry compliance check of fire safety rules.

NLP-assisted and hybrid code interpretation approaches attempt to address the bottleneck of manual rule authoring by extracting structured constraints directly from regulatory documents [22,23]. Early methods apply clause segmentation, dependency parsing, and semantic role labeling to identify regulated objects, parameter thresholds, and conditional logic [24,25]. More recent work incorporates semantic networks or knowledge graphs to bridge regulatory text with BIM semantics [26,27]. These studies show promise in reducing manual modeling effort, yet current NLP pipelines remain insufficient for handling the multi-layered semantics of fire safety codes, including tabular specifications, cross-referenced clauses, and context-dependent rule logic [28]. Moreover, existing approaches do not provide mechanisms to systematically map textual rule semantics to IFC geometry or user intent, limiting their applicability to geometry compliance check.

Overall, automated compliance check research has progressed from explicit, expert-authored rule engines with high interpretability but limited scalability to BIM query- and geometry-oriented methods with stronger engineering applicability and more recently to NLP- and LLM-assisted pipelines that aim to partially automate rule interpretation. Despite these advancements, existing approaches remain primarily oriented toward general or attribute-level regulatory conditions and fall short in addressing the multi-hop, context-dependent, and geometry-intensive requirements characteristic of fire safety rules. These gaps highlight the need for new frameworks capable of integrating semantic interpretation, geometric reasoning, and structured alignment into a cohesive and practicable workflow.

### 2.2. Large Language Models and Agent-based Reasoning: From General Frameworks to AEC Applications

LLMs have demonstrated remarkable capability in few-shot natural language interpretation and stepwise reasoning [29], driven by advances in prompting strategies that elicit multi-step cognitive processes. Chain-of-Thought (CoT) prompting enables LLMs to decompose complex tasks into intermediate reasoning steps, yielding strong performance in domains such as legal analysis, mathematical derivation, and structured decision-making [30]. Building on this idea, more expressive reasoning paradigms, such as Tree-of-Thought (ToT), which explores multiple reasoning branches simultaneously, and emerging graph-of-thought frameworks, which structure reasoning as graphs or executable programs, have further expanded LLMs' ability to navigate decision spaces, evaluate alternatives, and refine intermediate results [31,32]. Additional techniques such as self-consistency sampling [33], self-verification [34], and ReAct-style reasoning [35], which combines reasoning traces with external tool use [36,37], also demonstrate improved robustness for multi-step inference tasks [38,39].

On the other hand, recent research has developed multi-agent LLM frameworks such as CAMEL [40], AutoGen [41], MetaGPT [42], and other coordination-oriented architectures [43,44]. These systems assign specialized roles to different LLM agents, enabling task decomposition, iterative refinement, and collaborative problem-solving [45]. Multi-agent designs have proven effective for complex workflows requiring planning, feedback loops, and tool integration. Nevertheless, current multi-agent frameworks are generic in nature, which lack explicit grounding in domain-specific data structures and do not provide mechanisms to align agent reasoning with regulatory semantics or BIM geometry.

Emerging work in the AEC domain has begun to explore the use of large language models and agent-based techniques for tasks such as regulatory text summarization [46], BIM-oriented question answering [47,48], design assistance [49,50], construction planning support [51,52], and automated interpretation of technical documents [53,54]. These studies collectively illustrate the growing potential of LLM-driven intelligence to augment AEC workflows by improving information retrieval [55,56], supporting decision-making, and enhancing regulatory understanding [57]. However, existing applications typically focus on extracting or summarizing regulatory information, assisting with conversational BIM interfaces, or generating high-level design insights, but they do not provide mechanisms for multi-hop rule interpretation, multi-entity geometric dependency resolution, or cross-modal alignment in geometry compliance check applications. As a result, current LLM or agent-based approaches in AEC cannot support the structured, geometry-intensive reasoning processes required for fire-safety-related geometry compliance check. These gaps underscore the need for a domain-specific, alignment-driven method capable of integrating semantic interpretation, geometric reasoning, and interactive evaluation within a unified compliance workflow.

## 3. Method

This section delineates the underlying architecture and operational logic of SGR-BIM. Section 3.1 provides a high-level framework overview and a formal definition of the problem. Section 3.2 details the core mechanism of graph-driven cross-modal structured semantic alignment. Section 3.3 describes the implementation of the explainable and interactive interface designed for BIM-centered compliance check.

### 3.1. Framework Overview and Formal Problem Definition

The SGR-BIM framework is designed as a graph-driven multi-agent system that integrates user intent interpretation, geometry-related regulatory provision extraction, and BIM model interrogation into a unified workflow for automated geometry compliance check. As illustrated in Fig. 1, the system receives three forms of cross-modal input: user queries expressed in natural language, fire safety code provisions containing both narrative clauses and tabular specifications, and IFC-based BIM models that encode geometric and semantic building information. These inputs enter the system through the user interface, which also enables optional element selection and interactive exploration during the compliance process.

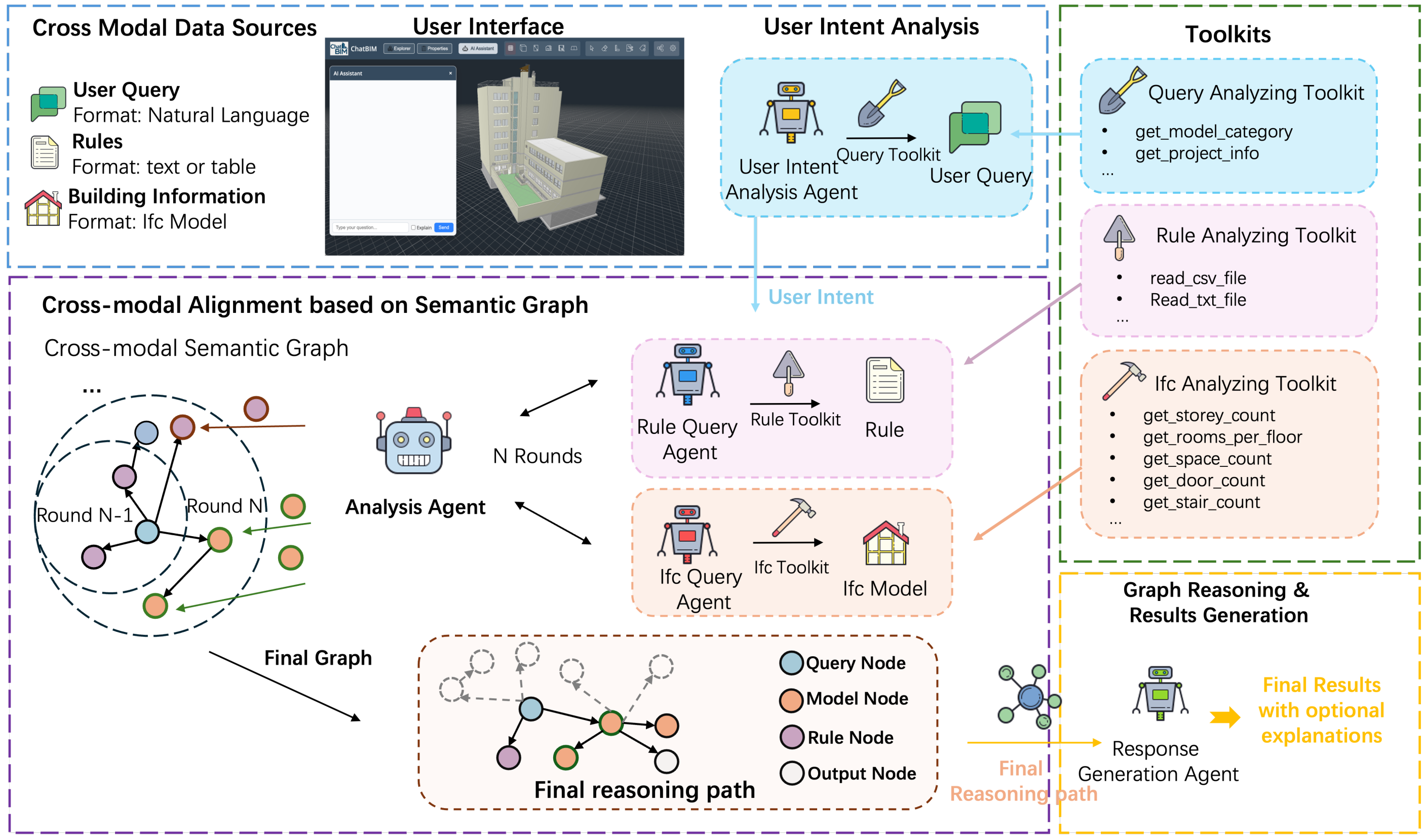

*Fig. 1 The overview of SGR-BIM framework, integrating cross-modal data sources, multi-agent coordination, and graph-driven semantic alignment.*

Within the framework, a set of specialized toolkits performs the initial interpretation and retrieval operations needed to access and structure these heterogeneous data sources. The query analysis toolkit offers functions for semantic parsing and project information extraction that support the agent's interpretation of user intent. The rule analysis toolkit exposes interfaces for locating, segmenting, and retrieving text or tabular entries from the fire safety code corpus. Similarly, the IFC analysis toolkit provides access to geometric attributes, spatial relationships, and entity metadata stored in the BIM model. These toolkits operate as retrieval utilities that supply the raw information required for downstream reasoning.

At the core of SGR-BIM lies the semantic graph that progressively aligns the user's intent, regulatory semantics, and BIM-derived geometric information. Through iterative expansion guided by the multi-agent workflow, the system constructs a unified embedding that integrates query nodes, rule nodes, and model nodes, linked through typed relations encoding reference dependencies, regulatory constraints, and spatial or topological relationships. This graph serves as the shared embedding space in which intermediate findings are accumulated, latent dependencies are resolved, and multi-hop reasoning paths required for geometry-intensive fire safety checks are constructed.

Once the semantic graph has been sufficiently populated, SGR-BIM performs graph-based reasoning to identify a final reasoning path that connects the user's query to the relevant regulatory requirements and model evidence. Based on this path, the system generates a result together with an optional explanation that presents the intermediate steps, supporting evidence, and derived geometric attributes. This enables transparent evaluation of geometry-related fire safety requirements and supports interactive, practitioner-oriented design review.

Building upon this high-level overview, we now formalize the core inputs and representations used in the framework. We denote a natural-language user query as

$$q \in Q \tag{1}$$

where $Q$ represents the space of admissible geometry-compliance questions. The regulatory corpus is represented as

$$R = \{r_1, r_2, \dots, r_m\} \tag{2}$$

where each $r_j$ is a narrative clause or tabular specification extracted from the fire safety code, potentially containing dimensional thresholds, applicability criteria, exceptions, and cross-references. The IFC BIM model is represented by the set

$$I = \{i_1, i_2, \dots, i_n\} \tag{3}$$

with each $i_k$ corresponding to an IFC entity (e.g., IfcDoor, IfcSpace, IfcStair), accompanied by geometric and semantic attributes retrieved through an IFC parsing toolset.

To support cross-modal alignment between these information sources, the SGR-BIM framework constructs a cross-modal semantic graph

$$G = (V, E) \tag{4}$$

in which nodes and edges encode semantic reference, regulatory constraints, and Ifc model spatial topology. The node set is defined as

$$V = V^Q \cup V^R \cup V^M \tag{5}$$

where a query node $v_q \in V^Q$ represents the structured interpretation of the user query. A rule node $v_r \in V^R$ represents a regulatory clause retrieved from $R$. A model node $v_m \in V^M$ represents an IFC entity from $I$ together with its geometric and semantic attributes. This representation embeds geometric properties directly within model nodes, ensuring that the graph remains compact while still supporting geometry-intensive reasoning.

Relations among nodes are represented using typed edges. To unify relation types, we define a relation space

$$\Lambda = \Lambda_{ref} \cup \Lambda_{reg} \cup \Lambda_{topo} \tag{6}$$

where $\Lambda_{ref}$ contains semantic reference relations linking query and model nodes (e.g., refers_to, targets). $\Lambda_{reg}$ contains regulatory constraint relations linking model nodes to rule nodes (e.g., constrained_by). $\Lambda_{topo}$ contains spatial and topological relations among model nodes (e.g., located_in, accesses, adjacent_to, connected_to). These relations are not defined as a closed or predefined set. Instead, they are dynamically inferred during the reasoning process based on available BIM data and contextual information. The edge set is therefore expressed as

$$E = \{(u, v, \tau) \mid u, v \in V, \tau \in \Lambda\} \tag{7}$$

The SGR-BIM framework employs five LLM-enabled agents whose roles differ in responsibility and granularity. The User Intent Analysis Agent produces the structured interpretation encoded in the query node,

$$f_{intent} : Q \longrightarrow (J_q, v_q) \tag{8}$$

where the natural-language query is transformed into a JSON-based intent specification $J_q$ that identifies the referenced building elements, relevant geometric attributes, and the implied regulatory scope. The agent then instantiates the query node $v_q$ and initializes the graph.

$$G_0 = (\{v_q\}, \emptyset) \tag{9}$$

Rule nodes and model nodes are added through subsequent iterations of the workflow. In the workflow, information retrieval is performed by two dedicated agents. The Rule Query Agent applies

$$f_{rule}: \Omega_{next} \longrightarrow R^* \tag{10}$$

where $\Omega_{next}$ is the instruction generated by the Analysis Agent, and $R^* \subseteq R$ denotes regulatory clauses relevant to the current information needs. The IFC Query Agent retrieves BIM entities through

$$f_{ifc}: \Omega_{next} \longrightarrow I^* \tag{11}$$

where $I^* \subseteq I$ denotes IFC entities and their geometric attributes required for the evaluation. Importantly, neither agent directly updates the graph. Instead, they supply the raw rule or model information needed for graph expansion. Graph construction and update are exclusively handled by the Analysis Agent, which orchestrates the workflow via

$$f_{analysis}: (G, q, H, T) \longrightarrow (S, \Omega_{next}, U) \tag{12}$$

where $H$ denotes the interaction history and $T$ denotes the evolving task specification. The output

$$S \in \{Comp, Rule, Model\} \tag{13}$$

indicates whether the graph is complete ($Comp$), requires additional regulatory information ($Rule$), or requires additional IFC model information ($Model$). The instruction

$$\Omega_{next} \in \Omega \tag{14}$$

describes what information the next retrieval agent must obtain. After retrieval, the Analysis Agent executes the graph-update operations encoded in

$$U = (V^{new}, E^{new}) \tag{15}$$

where each

$$V^{new} = \{(id, type, attrs)\} \tag{16}$$

represents a newly added or updated node with its attributes, and

$$E^{new} = \{(u, v, \tau) \mid u, v \in V, \tau \in \Lambda\} \tag{17}$$

epresents newly constructed edges. These operations directly extend the semantic graph and serve as the basis for the next iteration.

When $S = Comp$, the Response Generation Agent applies

$$f_{resp} : (G^*, H) \longrightarrow y \tag{18}$$

producing the final compliance determination supported by a final interpretable reasoning trace within the resulting semantic graph $G^*$.

Beginning with a single query node, the SGR-BIM framework repeatedly retrieves regulatory content and BIM data, updates the semantic graph, and aligns geometric, semantic, and contextual information until sufficient evidence is available to evaluate the query. This graph-driven formulation provides a unified mechanism for linking user intent, regulatory semantics, and BIM geometry, forming the foundation for the structured semantic alignment and reasoning procedures described in the following sections.

### 3.2. Graph-driven Cross-modal Structured Semantic Alignment

The core of the SGR-BIM framework is a graph-driven alignment mechanism that integrates user intent, regulatory semantics, and BIM-derived geometric information into a unified representation suitable for automated geometry compliance check. While user queries, code provisions, and IFC entities originate from heterogeneous modalities and operate at different levels of abstraction, geometry-related fire safety rules require these disparate sources to be jointly interpreted. The proposed semantic graph provides a shared, structured representation that enables multi-agent coordination, supports progressive fusion of cross-modal information, and facilitates explainable reasoning over geometry-intensive regulatory requirements. This section introduces the graph-driven cross-modal alignment mechanism supported by a multi-agent iterative coordination workflow. Fig 1 and Fig 2 jointly illustrate the conceptual architecture and the operational process behind this mechanism.

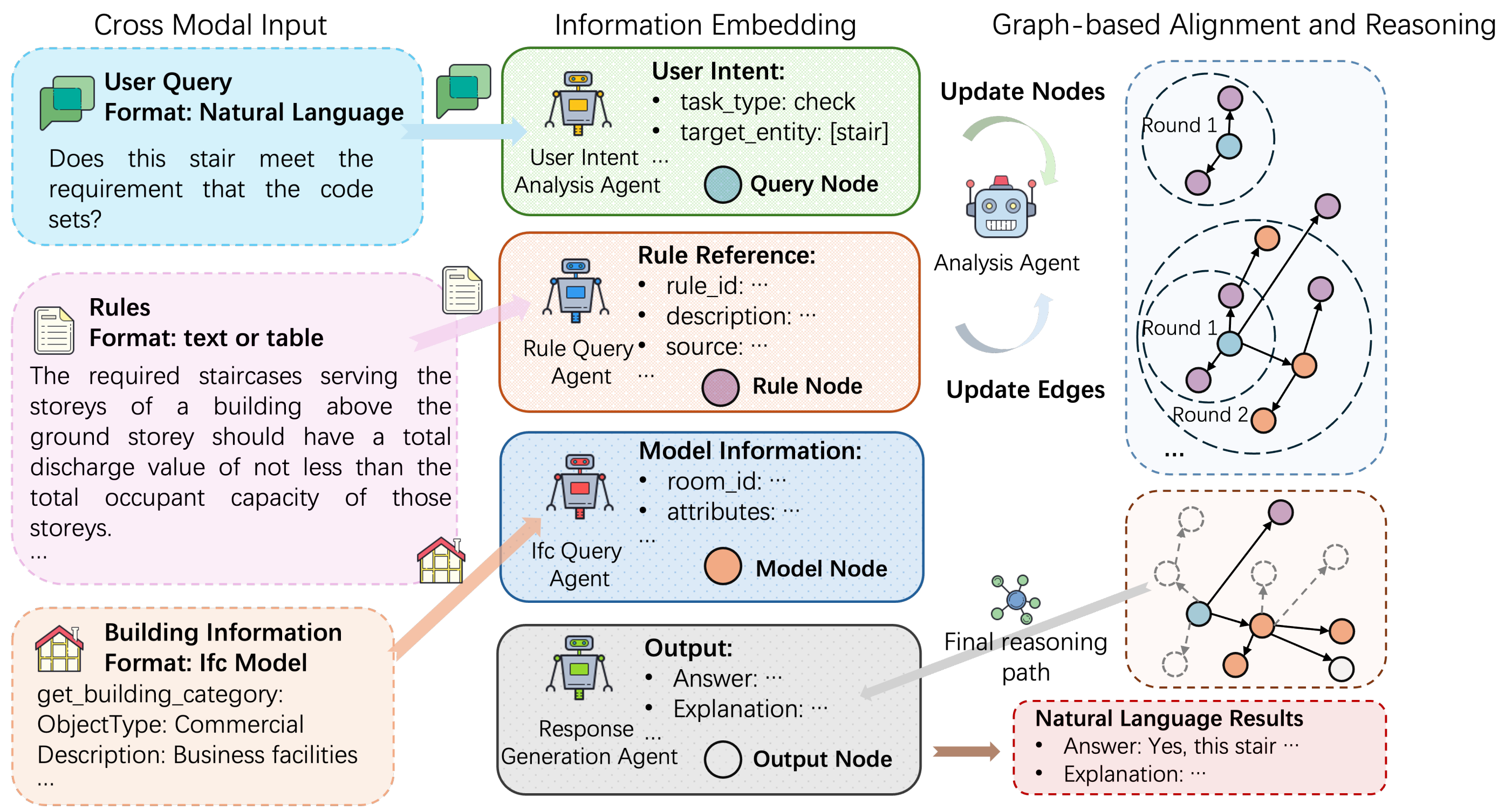

*Fig. 2 Cross-modal alignment pipeline of SGR-BIM, including information embedding, graph construction, and graph-based reasoning.*

To bridge the semantic gaps across modalities, SGR-BIM embeds information from user intent, regulatory clauses, and BIM model entities into a unified semantic graph. Beginning with the natural-language query $q$, the User Intent Analysis Agent applies

$$(J_q, v_q) = f_{intent}(q) \tag{19}$$

where $J_q$ denotes the structured intent specification and $v_q \in V^Q$ is the corresponding query node that initializes the graph $G_0 = (\{v_q\}, \emptyset)$. Subsequent information is progressively integrated through a combination of retrieval operations and analysis-driven graph updates. Regulatory fragments returned by the Rule Query Agent,

$$R^* = f_{rule}(\Omega_{next}) \tag{20}$$

and IFC-derived geometric content returned by the IFC Query Agent,

$$I^* = f_{ifc}(\Omega_{next}) \tag{21}$$

are not inserted into the graph directly. Instead, the Analysis Agent transforms these raw retrieval results into new rule nodes $v_r$, model nodes $v_m$, and typed edges that reflect semantic or spatial relationships defined in the regulatory text or encoded in the BIM model. Through this information embedding process, SGR-BIM generates a structured, modality-aware graph representation in which each node encapsulates content relevant to a specific modality, and each edge indicates a reference, constraint, or topological relationship between entities.

The expansion and refinement of this semantic graph is driven by an iterative coordination loop orchestrated by the Analysis Agent. At each iteration, the Analysis Agent examines the current graph state $G$, together with the user's query $q$, the accumulated interaction history $H$ and task specification $T$, to determine whether the information contained in the graph is sufficient for evaluating the compliance question. Formally, the agent applies

$$(S, \Omega_{next}, U) = f_{analysis}(G, q, H, T) \tag{22}$$

where $S \in \{Comp, Rule, Model\}$ indicates whether the alignment is complete ($Comp$), requires additional regulatory information ($Rule$), or requires additional geometric information from the BIM model ($Model$).

This iterative process continues until all required cross-modal dependencies, including rule thresholds, exception conditions, spatial relationships, or latent entity linkages, have been embedded into the graph. Once the alignment converges ($S = Comp$), the resulting semantic graph $G^*$ encodes a complete cross-modal representation of the compliance problem. The Response Generation Agent then performs graph-based reasoning

$$y = f_{resp}(G^*, H) \tag{23}$$

identifying a reasoning path

$$\pi = \{v_q \longrightarrow v_{m_1} \longrightarrow \cdots \longrightarrow v_{r_k}\} \tag{24}$$

that links the user query to the relevant BIM entities and regulatory clauses. Because each step of this chain is explicitly represented as a node or edge within the graph, the reasoning process is inherently transparent. The system can therefore provide interpretable explanations, reveal intermediate assumptions, and support interactive exploration of compliance outcomes, capabilities that are difficult to achieve with rule-based compliance check approaches.

Algorithm 1 formalizes this workflow, detailing the interaction among agents, the retrieval of cross-modal information, and the recursive integration of newly acquired content. Together with Fig. 2, the algorithm highlights how SGR-BIM incrementally transforms heterogeneous inputs into a tightly aligned semantic structure capable of supporting graph-based geometry compliance reasoning. Through this graph-driven cross-modal alignment mechanism, SGR-BIM offers a robust and explainable foundation for automated geometry compliance check of fire safety codes.

**Algorithm 1**. Graph-driven multi-agent workflow for cross-modal semantic alignment

**Input:** User query $q$, regulatory corpus $R$, IFC model $I$

**Output:** Aligned semantic graph $G^*$, compliance result $y$

1: $(J_q, v_q) \longleftarrow f_{intent}(q)$

2: Initialize $G_0 \longleftarrow (\{v_q\}, \emptyset)$

3: Initialize history $H$ and task specification $T$

4: **while** $S \neq Comp$ **do**

5: $(S, \Omega_{next}, U) \longleftarrow f_{analysis}(G, q, H, T)$

6: if $S = Rule$ then

7: $R^* \longleftarrow f_{rule}(\Omega_{next})$

8: Update $H$ and $T$ with $R^*$

9: if $S = Model$ then

10: $I^* \longleftarrow f_{ifc}(\Omega_{next})$

11: Update $H$ and $T$ with $I^*$

12: Apply update $U$ to $G$

13: **end while**

14: $y \longleftarrow f_{resp}(G^*, H)$

15: return $(G^*, y)$

### 3.3. Explainable and Interactive Geometry Compliance Check Interface

Automated compliance check is most effective when the results are not only correct but also transparent and easily interpretable by practitioners. In the context of fire safety regulations, where geometric requirements often involve multi-step reasoning, latent dependencies among building entities, and conditional logic embedded in the code, explainability and interactivity are essential for enabling designers, engineers, and reviewers to understand and verify the basis of compliance results. Traditional automated compliance systems typically operate as closed pipelines: users provide design models and receive pass–fail outcomes, with limited visibility into intermediate reasoning or the specific geometric and regulatory evidence used to support those decisions. Such opacity can reduce trust, hinder adoption in practice, and limit the system's usefulness during iterative design refinement.

To address these limitations, SGR-BIM incorporates an explainable and interactive user interface that exposes the underlying reasoning process of the graph-driven multi-agent framework. The interface enables users to visualize BIM geometry, inspect IFC entities, explore the cross-modal semantic graph, and interact with results at the level of individual rules, geometric attributes, and reasoning steps. By coupling the alignment mechanism described in Section 3.2 with an interactive visualization layer, SGR-BIM provides a transparent compliance environment in which users can both query and interrogate the system's conclusions.

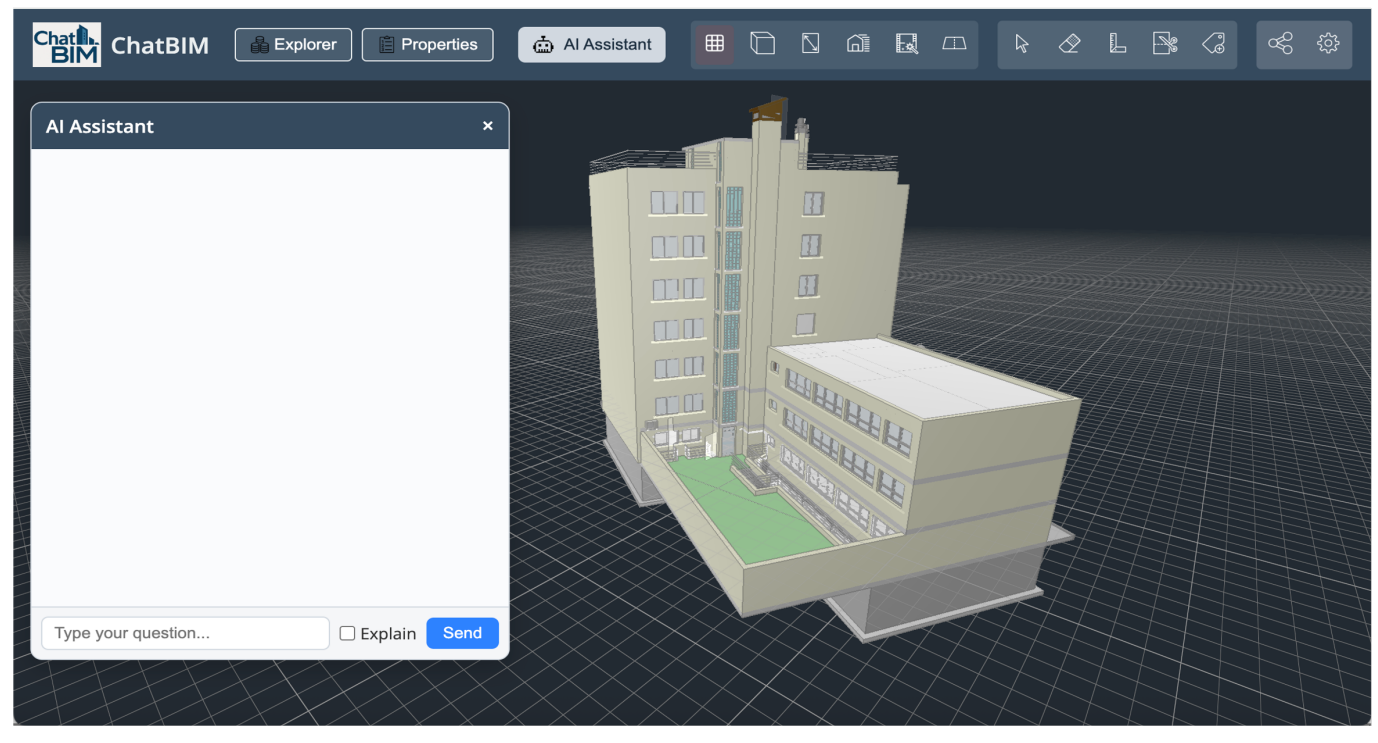

(a) Overview of the SGR-BIM user interface

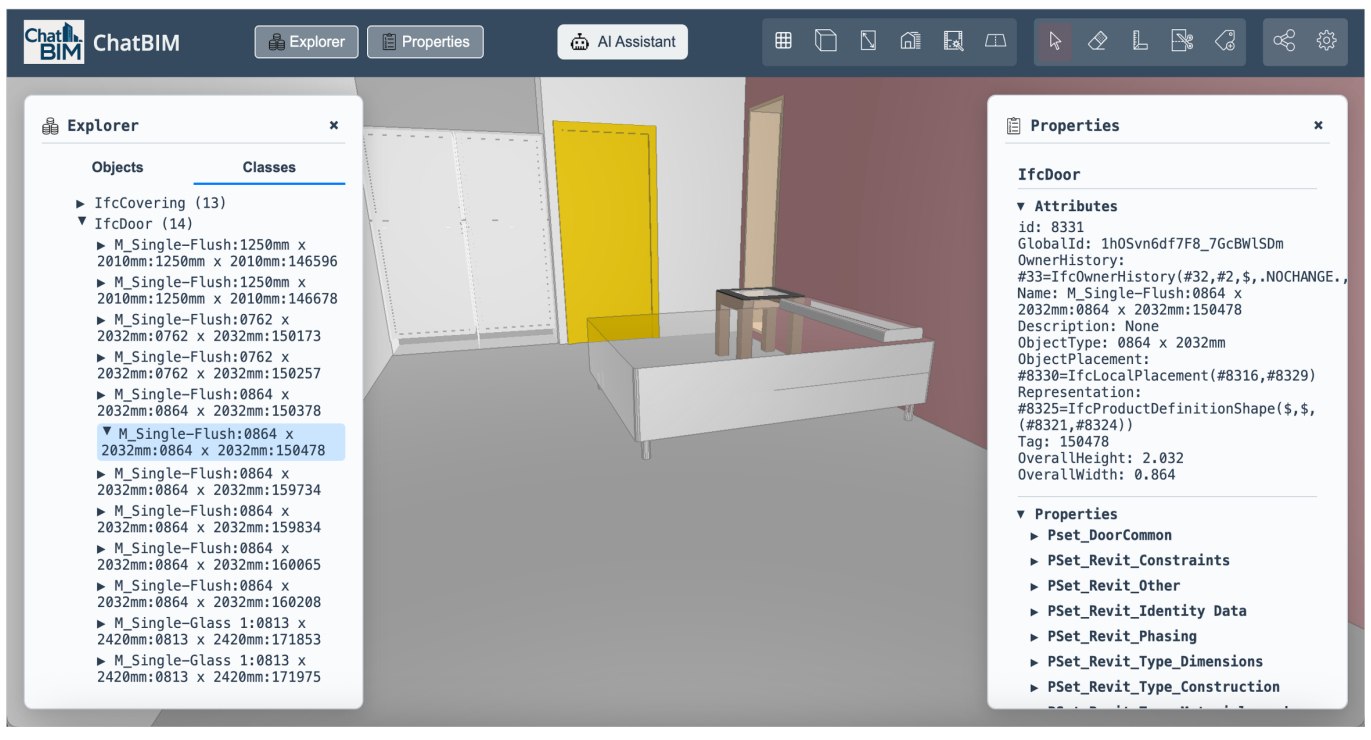

(b) Interactive entity inspection

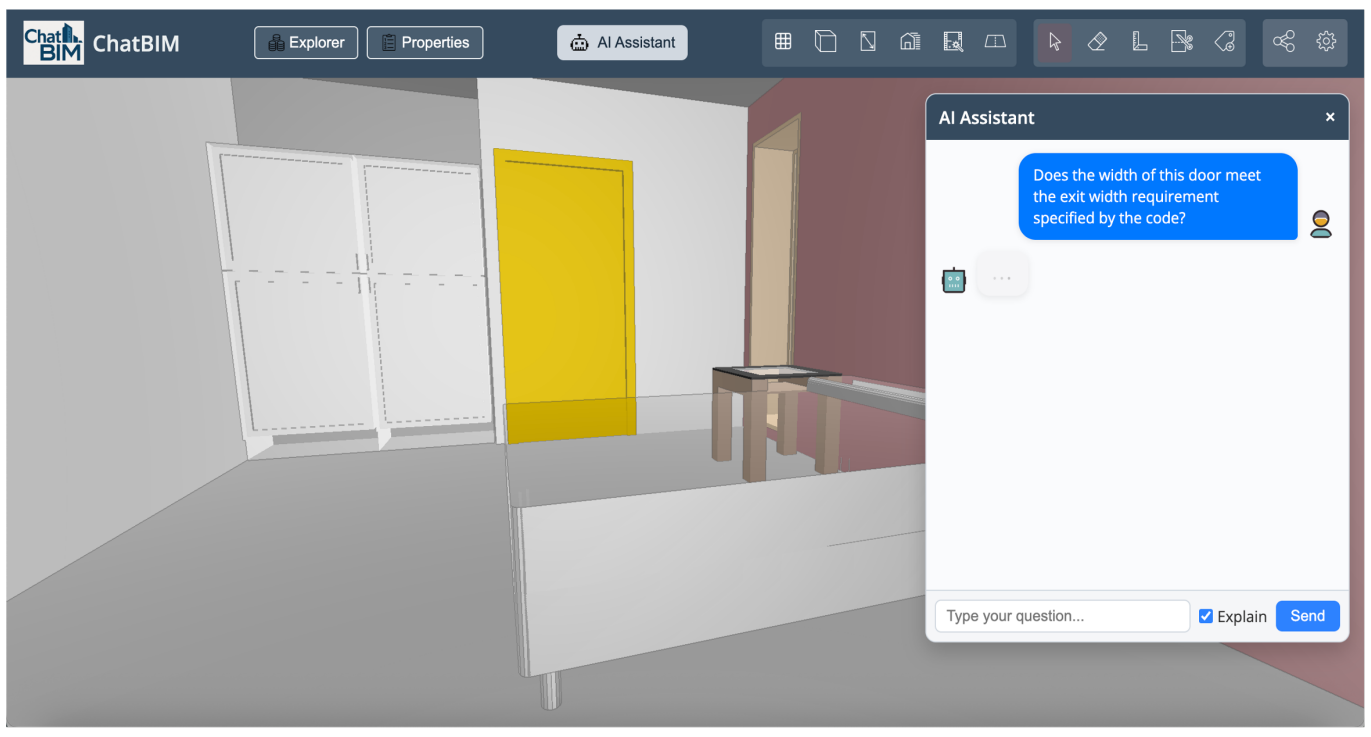

(c) Referential geometry compliance check query

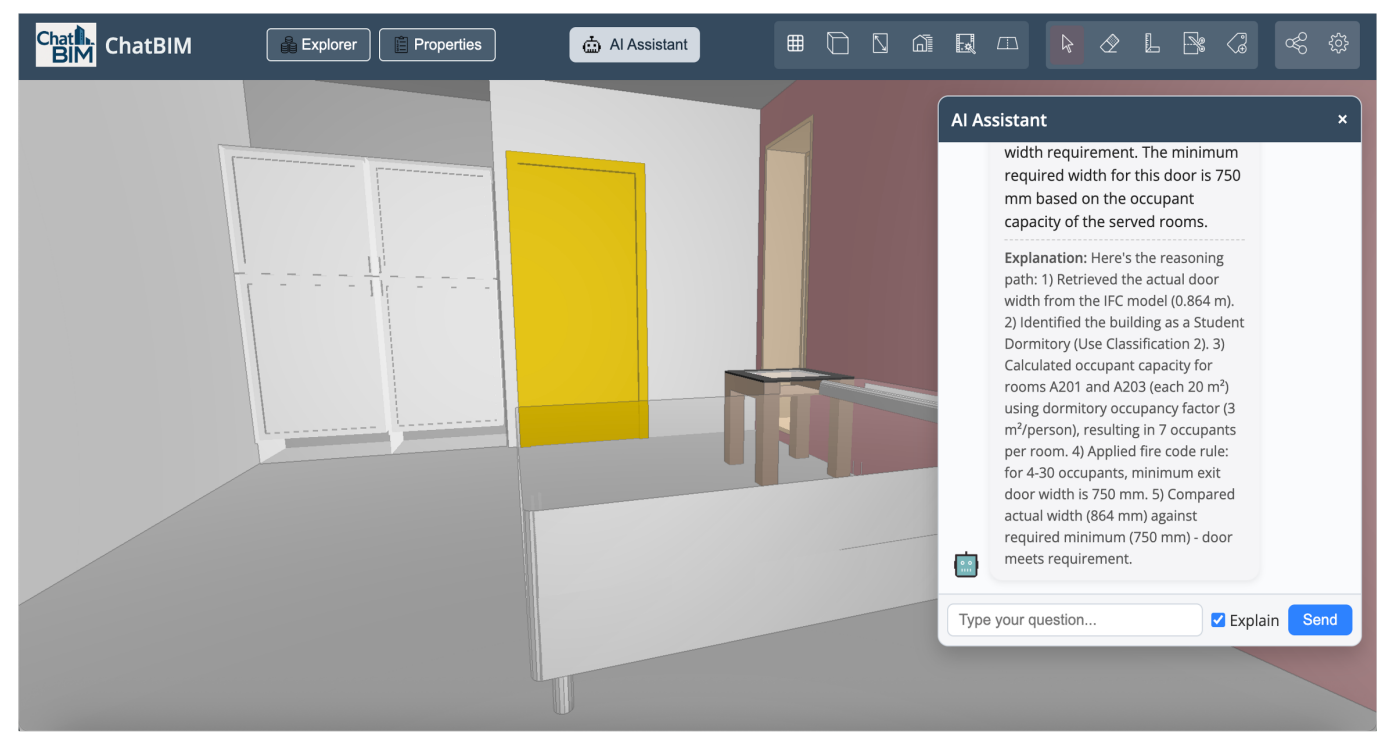

(d) Explainable compliance result

*Fig. 3 Interactive interface of SGR-BIM.*

Fig 3(a) presents an overview of the SGR-BIM interface. The system displays the user's BIM model in an interactive 3D viewport that supports entity selection, highlighting, and geometric inspection. Geometry compliance check is presented as one of the system's core functions and is accessed through a dedicated chat window that allows users to enter natural-language queries. The interface includes a checkbox for enabling explanation mode, allowing practitioners to request fully interpretable reasoning paths when needed. This blended design, which combines direct model interaction with conversational querying, enables flexible workflows in which users can either inspect geometry visually or request automated evaluation of specific code-related questions.

To support entity-level understanding, the interface allows users to click on components within the BIM model, such as doors, rooms, or stair elements, to access detailed information about their geometric and semantic attributes. As shown in Fig 3(b), selected elements are highlighted in the viewport, and the corresponding attribute panel displays relevant IFC data, including attributes, dimensions, and spatial relationships. This feature supports interactive exploration of model content and facilitates contextual understanding during compliance check, especially in cases where geometric dependencies span multiple building entities.

Beyond passive inspection, the interface supports referential queries that ground user intent in the selected BIM entities. As illustrated in Fig 3(c), after selecting an object in the model, the user may pose a query such as "Does the width of this door meet the exit width requirement specified by the code?". The system automatically interprets the referring expression "this door", aligns it with the selected IFC entity, and invokes the graph-driven reasoning pipeline to evaluate the relevant geometric rule chain. This capability demonstrates the integration between model interaction and semantic alignment: user selections directly influence the construction of the semantic graph and guide the agents in retrieving the appropriate regulatory and geometric information.

Finally, when explanation mode is enabled, SGR-BIM provides an interpretable reasoning trace alongside the compliance check outcome. Fig 3(d) shows an example in which the system not only returns a result but also presents a natural-language explanation detailing the geometric attributes considered, the specific regulatory clauses applied, and the intermediate deductions made during reasoning. This explanation is derived from the reasoning path extracted from the aligned semantic graph and therefore reflect the actual sequence of rule dependencies and IFC evidence used in evaluation. By exposing this chain of reasoning, the interface supports transparency, facilitates trust, and enables practitioners to validate or challenge the underlying assumptions during design review.

Through these explainable and interactive interface features, SGR-BIM enhances the practical usability of the underlying multi-agent alignment framework. By enabling users to explore BIM geometry, inspect regulatory evidence, and interrogate reasoning outcomes, the interface bridges the gap between automated reasoning and decision-making, providing a supportive environment for reliable, interpretable, and user-driven geometry compliance check of fire safety codes.

## 4. Experiment

This section describes the experimental validation of SGR-BIM using a diverse set of fire safety regulatory queries. Section 4.1 outlines the construction of the benchmark dataset and the definition of evaluation metrics, while Section 4.2 details the experimental setup and baseline configurations. The quantitative performance results and a diagnostic error analysis are presented in Section 4.3 and Section 4.4, respectively. The section concludes with Section 4.5, which provides a discussion of the framework's reasoning characteristics and practical applicability.

### 4.1. Question Dataset and Evaluation Metrics

To systematically evaluate the robustness of SGR-BIM in handling complex, geometry-intensive regulatory constraints, we constructed a specialized benchmark dataset consisting of 679 questions derived from practical BIM-based design review scenarios. Fire safety regulations were selected as the representative validation domain for this study, as they impose some of the most stringent and intricate spatial requirements in building design (e.g., multi-hop travel distances, topology-dependent separations), thereby serving as an ideal rigorous testbed for geometric reasoning capabilities. The dataset covers two major phases of practical workflows, design-stage inquiries and check-stage evaluations, and organizes questions into four semantically coherent categories that reflect generic information needs in spatial compliance reasoning, instantiated here through specific fire safety provisions.

The dataset is constructed based on five IFC models covering diverse building types, including two multi-storey residential apartment buildings, one dormitory, one commercial office building, and one industrial factory. These models were selected to ensure diversity in building use classification, spatial configuration, and scale, thereby reflecting a broad range of realistic design scenarios encountered in practice. The initial models were obtained from publicly available sources and subsequently refined by a research team with BIM expertise, consisting of three members with over one year of BIM experience, including one practitioner with extensive industry experience, to ensure consistency, completeness, and suitability for compliance checking tasks.

The question set was developed through a semi-structured process combining domain expertise and LLM-assisted expansion. First, domain experts analyzed each IFC model to identify representative compliance scenarios and manually formulated a set of seed questions targeting key regulatory and geometric reasoning challenges. These seed questions were then expanded using large language models to improve coverage and diversity. Finally, all generated questions were manually reviewed and validated by the expert team to ensure correctness, relevance, and alignment with realistic compliance workflows.

The ground truth answers were entirely produced by domain experts through detailed analysis of both the regulatory provisions and the corresponding BIM models. This process ensures that each answer is consistent with regulatory requirements and accurately reflects the underlying geometric and semantic conditions of the model, thereby providing a reliable benchmark for evaluating automated compliance checking performance.

Design-stage questions focus on understanding regulatory requirements and deriving design targets prior to model development. They include:

- Rule questions, which request specific geometry-related provisions from the fire safety code (e.g., maximum compartment areas).
- Rule requirement questions, which query the required geometric properties for a specific building element (e.g., the clear width required for a particular door).

Check-stage questions evaluate as-built or as-modeled BIM geometry against regulatory constraints. They include:

- Model questions, which retrieve geometric or spatial information from the IFC model (e.g., floor level, dimensions, or adjacency).
- Check questions, which assess whether an IFC element satisfies the applicable geometry-dependent fire safety requirements.

Table 1 Overview of the Question Dataset Across Stages and Categories

| Stage | Category | Number | Multiple Rule Sources | Geometric Computing | Relational Dependencies | Example |
|---|---|---|---|---|---|---|
| Design-stage | Rule | 113 | 79.65% | / | / | What is the maximum allowable fire compartment area for commercial buildings? |
| | Rule Requirement | 189 | / | 53.97% | 10.05% | What clear width does the code actually require for this door? |
| Check-stage | Model | 240 | / | 57.92% | 40.00% | On which floor is this room located? |
| | Check | 137 | / | 85.40% | 44.52% | Does this stair comply with the code requirements? |
| Overall | / | 679 | / | / | / | / |

Table 1 summarizes the dataset composition, including representative examples from each category. Specifically, the technical difficulty of the benchmark is characterized by three functional dimensions of complexity. First, for the Rule category, complexity is defined by the number of regulatory sources required to formulate a correct response. 79.65% of rule-related queries are classified as high-complexity, as they necessitate the simultaneous interpretation of two or more distinct regulatory clauses.

Second, for the Rule Requirement, Model, and Check categories, complexity is driven by Geometric Computing and Relational Dependencies. Geometric Computing involving spatial quantitative analysis, accounts for 53.97% to 85.40% of the queries in these categories. Relational Dependencies capture the necessity of navigating topological or logical links between two or more interlinked IFC entities.

Finally, the dataset incorporates high-level logical operations to simulate realistic professional workflows, including table lookup operations integrated across all categories. Furthermore, each query category for every building model contains 1–2 conditional exceptions, testing the framework's ability to handle edge cases where standard rules may be superseded by specific design contexts.

To assess model performance across this dataset, four evaluation dimensions were defined, Accuracy, Coherence, Relevance, and Explainability & Reasoning, capturing complementary aspects of response quality. Accuracy serves as the primary quantitative metric, reflecting factual correctness. To provide a more granular and rigorous assessment beyond a simple binary judgment, Accuracy is evaluated on a three-tier scale of 0, 0.5, and 1. A score of 1 (Fully Correct) is awarded only when the response delivers the correct final conclusion while simultaneously identifying all governing regulatory variables and accounting for situational boundary conditions defined in the code. This ensures that the framework demonstrates a complete understanding of the underlying regulatory logic and situational variations, rather than merely retrieving isolated data points.

A score of 0.5 (Partially Correct) is assigned when the system identifies the correct underlying regulatory mechanisms or relevant IFC entities but reaches an imprecise conclusion, or fails to address all necessary thresholds and specific edge cases required for a comprehensive compliance check. This intermediate score allows for the differentiation between responses that are logically grounded but incomplete and those that are fundamentally flawed. Conversely, a score of 0 (Incorrect) is given when both the final conclusion and the underlying reasoning path are found to be erroneous or based on hallucinated information. The remaining three metrics, Coherence, Relevance, and Explainability & Reasoning, provide qualitative insights into the clarity, completeness, and transparency of the generated responses. These qualitative dimensions are essential for ensuring that the system's outputs are not only accurate but also practical and trustworthy within regulatory workflows. Table 2 summarizes these metrics and provides their formal scoring criteria.

Table 2 Evaluation Metrics and Scoring Criteria

| Dimension | Definition | Key Judgment Criteria | Scoring Scale |
|---|---|---|---|
| **Accuracy** | Whether the answer is factually correct and aligned with fire code context. | - Fully correct (1)<br>- Partially correct (0.5)<br>- Incorrect (0) | 0-1 |
| **Coherence** | Degree to which the answer fully covers all relevant aspects, cases, or conditions implied by the question. | - No unnatural references (*this rule*, *this table*) without context<br>- No long GUID/random strings<br>- Units properly formatted, readable, no encoding errors<br>- If multiple cases apply, examples are provided<br>- Overall fluency and readability | 0–1 |
| **Relevance** | Whether the answer directly addresses the user's question. | - Covers the core point(s) of the question<br>- No digressions or irrelevant content<br>- Provides a direct response, not just background | 0–1 |
| **Explainability & Reasoning** | Extent to which the answer provides a clear reasoning chain. | - Shows reasoning, not only conclusion<br>- Logical chain is transparent and easy to follow<br>- Edge cases / exceptions are explained if relevant | 0–1 |

Overall, the dataset provides a comprehensive and hierarchically structured benchmark that reflects the full spectrum of geometry-intensive information needs encountered in BIM-based fire safety workflows. By combining design-stage regulatory inquiries with check-stage model evaluations, it enables a rigorous assessment of both rule interpretation and geometry-grounded reasoning. The evaluation metrics complement this design by capturing not only factual correctness but also the clarity, relevance, and transparency required for reliable compliance support. This dataset–metric evaluation system establishes the foundation for the quantitative analyses presented in the following sections.

### 4.2. Experimental Setup and Baseline Models

To rigorously evaluate performance across the question categories and metrics introduced in Section 4.1, we configure a controlled experimental environment in which all models, both our proposed SGR-BIM framework and the baseline systems, operate under consistent inference settings, tool access, and evaluation protocols. All experiments were conducted using large language model (LLM) backends under identical inference configurations to ensure fair comparison. Unless otherwise stated, the default backend for SGR-BIM and all baselines was DeepSeek-3.2, with a temperature of 0 to eliminate sampling randomness and enforce deterministic reasoning behavior. This configuration allows the evaluation to focus on the structural differences across reasoning strategies, particularly the contributions of tool usage, multi-agent coordination, and graph-driven cross-modal alignment, rather than stochastic generation variance.

Under this deterministic configuration, all experiments were conducted in a single run. Preliminary small-scale tests indicated that model outputs remain stable when the temperature is set to 0, suggesting minimal variance introduced by the generation process. Given the relatively large size of the benchmark (679 questions) and the involvement of human verification for metrics, repeating the full experiment multiple times would incur substantial computational and annotation costs. Therefore, a single-run evaluation strategy was adopted in this study.

In our study, the information retrieval for regulatory codes and BIM model is implemented via dedicated functions that are wrapped as tools for the corresponding agents. These functions directly access structured information from pre-processed data sources and, through prompt design (as summarized in Appendix A), convert it into a format interpretable by agents and incorporated into the semantic graph for context-aware reasoning. This approach allows the agents to perform multi-hop reasoning by iteratively invoking the retrieval tools as needed, while keeping the implementation deterministic and straightforward.

As the retrieval tools operate deterministically, the effective success of information access primarily depends on the agent correctly calling the functions with appropriate parameters and timing, which in our controlled experimental setup is highly reliable. One limitation of this approach is that the underlying data sources should be pre-processed and correctly formatted to ensure accurate information extraction. However, as the focus of this study is on cross-modal reasoning and semantic alignment, we intentionally do not integrate more complex retrieval frameworks [46,58], such as RAG or SPARQL, which are commonly applied in information retrieval for multimodal building regulations or IFC models.

To isolate these contributions, we designed a set of baseline models that span a spectrum of reasoning capabilities, from single-agent chain-of-thought reasoning to well-established multi-agent collaboration frameworks. The Basic-tool Agent represents the lowest level of reasoning support, allowing an LLM to access only basic tools for retrieving rule text or BIM attributes. It does not incorporate any structured alignment strategy and thus reflects purely sequential reasoning driven by textual context. Building on this, the Enhanced-tool Agent grants access to the full domain-specific toolset used by SGR-BIM, including rule retriever and IFC geometric query utilities. This baseline isolates the benefit of tool availability from the additional gains that may arise from coordination among agents or cross-modal semantic alignment.

Beyond single-agent systems, we include three widely recognized multi-agent frameworks, CAMEL, MetaGPT, and AutoGen, to evaluate whether general-purpose collaborative reasoning mechanisms can address the unique demands of geometry fire safety compliance check tasks. CAMEL employs two conversational agents in role-specific dialogue, enabling collaborative task decomposition but lacking any explicit representation of relationships between regulatory provisions and BIM geometry. MetaGPT follows a workflow-driven design, organizing multiple specialized agents (e.g., Planner, Reviewer) into a predefined pipeline that sequentially processes subtasks. AutoGen adopts a more flexible conversational architecture, where a group of agents engage in iterative negotiation and self-correction while performing the task. Although these frameworks embody increasingly sophisticated coordination strategies, none provides explicit semantic alignment and reasoning among regulatory text, model entities, and user intent.

By contrast, SGR-BIM integrates multi-agent collaboration with a graph-driven cross-modal semantic alignment mechanism that explicitly links regulatory clauses, IFC entities, and structured user intent. This alignment allows agents to operate over a shared semantic graph, enabling multi-hop reasoning, latent entity linkage resolution, and transparent compliance check. Table 3 summarizes the comparative characteristics of each baseline considered in the evaluation.

Table 3 Characteristics of Baseline Models and Our Model

| Model | Multi-agent | Cross-modal Data Alignment & Reasoning |
|---|---|---|
| **Basic-tool Agent** | No | Chain-of-thought reasoning over data inputs using basic tools, without explicit cross-modal alignment |
| **Enhanced-tool Agent** | No | Chain-of-thought reasoning with complete domain-specific toolset usage, without explicit cross-modal alignment |
| **CAMEL Baseline** | Yes (2 agents: auditor & assistant) | Dialogue-based collaborative reasoning between agents, without explicit cross-modal alignment |
| **MetaGPT Baseline** | Yes (4 specialized roles: planner, analyzer, executor, reviewer) | Workflow-driven role coordination with sequential task passing; no explicit cross-modal alignment |
| **AutoGen Baseline** | Yes (3 agents: auditor, reasoner, reviewer) | Dynamic conversational reasoning with self-corrective feedback; no explicit structured alignment |
| **SGR-BIM** | Yes (5 agents: intent, model, rule, analysis, response) | Graph-based cross-modal alignment integrating rule semantics, IFC geometry representations, and user intent |

These baselines collectively provide a tiered comparison spanning single-agent reasoning, tool-augmented processing, and multi-agent collaboration without structured alignment. This design allows us to systematically disentangle the contributions of tool integration, multi-agent coordination, and graph-based cross-modal alignment. Section 4.3 presents the quantitative performance results across all baselines.

### 4.3. Quantitative Performance

This section presents the comprehensive quantitative results derived from the comparative experiments to evaluate the effectiveness of the SGR-BIM framework.

#### 4.3.1. Overall Accuracy Performance

Building on the experimental setup and baseline models introduced in Section 4.2, we first evaluate overall accuracy performance across the 679 geometry-related fire safety queries. Table 4 presents the correct, partial-correct, and incorrect response rates for each model across all stages and categories, together with the corresponding average accuracy scores.

Table 4 Overall performance across all stages and categories for baseline models and SGR-BIM.

| Stage | Category | Baseline | Number | Correct(%) | Partial(%) | Wrong(%) | Avg Score |
|---|---|---|---|---|---|---|---|
| Design-stage | Rule | Basic-tool agent | 113 | 70.8 | 15.0 | 14.2 | **78.3** |
| | | Enhanced-tool agent | | 82.3 | 8.8 | 8.8 | **86.7** |
| | | Camel Baseline | | 81.4 | 5.3 | 13.3 | **84.1** |
| | | metaGPT Baseline | | 73.5 | 15.0 | 11.5 | **81.0** |
| | | Autogen Baseline | | 77.0 | 5.3 | 17.7 | **79.6** |
| | | **SGR-BIM** | | **87.6** | **6.2** | **6.2** | **90.7** |
| | Rule Requirement | Basic-tool agent | 189 | 0 | 16.4 | 83.6 | **8.2** |
| | | Enhanced-tool agent | | 65.6 | 9.5 | 24.9 | **70.4** |
| | | Camel Baseline | | 60.8 | 10.6 | 28.6 | **66.1** |
| | | metaGPT Baseline | | 27.5 | 45.5 | 27.0 | **50.3** |
| | | Autogen Baseline | | 50.3 | 19.0 | 30.7 | **59.8** |
| | | **SGR-BIM** | | **72.0** | **13.2** | **14.8** | **78.6** |
| Check-stage | Model | Basic-tool agent | 240 | 5.4 | 5.8 | 88.8 | **8.3** |
| | | Enhanced-tool agent | | 70.4 | 12.1 | 17.5 | **76.5** |
| | | Camel Baseline | | 76.2 | 1.7 | 22.1 | **77.1** |

| | | metaGPT Baseline | | 69.6 | 5.0 | 25.4 | **72.1** |
|---|---|---|---|---|---|---|---|
| | | Autogen Baseline | | 75.0 | 9.2 | 15.8 | **79.6** |
| | | **SGR-BIM** | | **85.8** | **4.2** | **10** | **87.9** |
| | Check | Basic-tool agent | 137 | 0 | 8.0 | 92.0 | **4.0** |
| | | Enhanced-tool agent | | 66.4 | 12.4 | 21.2 | **72.6** |
| | | Camel Baseline | | 57.7 | 19.0 | 23.4 | **67.2** |
| | | metaGPT Baseline | | 48.2 | 22.6 | 29.2 | **59.5** |
| | | Autogen Baseline | | 57.7 | 5.1 | 37.2 | **60.2** |
| | | **SGR-BIM** | | **75.2** | **10.9** | **13.9** | **80.7** |
| / | **Overall** | Basic-tool agent | 679 | 13.7 | 10.8 | 75.6 | **19.1** |
| | | Enhanced-tool agent | | 70.3 | 10.9 | 18.9 | **75.7** |
| | | Camel Baseline | | 69.1 | 8.2 | 22.7 | **73.2** |
| | | metaGPT Baseline | | 54.2 | 21.5 | 24.3 | **64.9** |
| | | Autogen Baseline | | 64.9 | 10.5 | 24.6 | **70.2** |
| | | **SGR-BIM** | | **80.1** | **8.4** | **11.5** | **84.3** |

Across the full dataset, SGR-BIM achieves the highest overall accuracy of 84.3%, outperforming the generic multi-agent baselines by approximately 10%. On the other hand, the framework demonstrates a measurable 8.6% improvement over the Enhanced-tool Agent (75.7%). This performance margin suggests that while providing an LLM with specialized tools is sufficient for resolving a large portion of standard queries, the graph-driven multi-agent coordination becomes essential for complex reasoning. Specifically, the framework shows robustness in tasks where isolated tool invocations are insufficient to resolve latent spatial dependencies or multi-hop semantic alignments, proving that the added knowledge graph alignment yields tangible benefits for high-difficulty geometric compliance check tasks.

Performance trends at the stage level further highlight this difference. In the design stage, SGR-BIM achieves 90.7% accuracy for rule queries, exceeding all baselines, and reaches 78.6% in the rule requirement category. The latter represents a substantial improvement over CAMEL (66.1%), AutoGen (59.8%), and MetaGPT (50.3%), indicating that deriving element-specific requirements from regulatory logic particularly benefits from structured semantic alignment.

A similar pattern appears in the check stage, where SGR-BIM attains 87.9% accuracy for model queries and 80.7% for the more challenging geometry compliance check tasks. In both categories, SGR-BIM clearly outperforms all baselines, including the Enhanced-tool Agent (76.5% and 72.6%, respectively) and general-purpose multi-agent models. The largest gains occur in the compliance check category, where reasoning requires linking multiple IFC entities, interpreting conditional rule chains, and integrating geometric attributes, a process that general multi-agent systems struggle to handle without explicit alignment mechanisms.

Table 5 Tukey HSD pairwise comparison results among different methods.

| **Method A** | **Method B** | **Mean Difference (A − B)** | **p-value** | **Significant or not** |
|---|---|---|---|---|
| Enhanced-tool agent | Basic-tool agent | +0.5663 | <0.0001 | √ |
| Camel Baseline | Enhanced-tool agent | -0.0250 | 0.8502 | × |
| metaGPT Baseline | Enhanced-tool agent | -0.1075 | <0.0001 | √ |
| Autogen Baseline | Enhanced-tool agent | -0.0552 | 0.1010 | × |
| SGR-BIM | Basic-tool agent | +0.6524 | <0.0001 | √ |
| SGR-BIM | Enhanced-tool agent | +0.0862 | 0.0008 | √ |
| SGR-BIM | Camel Baseline | +0.1112 | <0.0001 | √ |
| SGR-BIM | metaGPT Baseline | +0.1937 | <0.0001 | √ |
| SGR-BIM | Autogen Baseline | +0.1414 | <0.0001 | √ |

To further examine the performance differences among methods, we conducted a Tukey HSD pairwise comparison based on per-question results across the full dataset. The results are summarized in Table 5. The analysis shows that SGR-BIM consistently outperforms all baseline methods with statistically significant differences. In particular, the performance gains over the Basic-tool Agent (+0.6524) and MetaGPT (+0.1937) are substantial, while improvements over other multi-agent frameworks such as AutoGen (+0.1414) and CAMEL (+0.1112) remain consistently significant. When compared to the Enhanced-tool Agent, SGR-BIM still achieves a statistically significant improvement (+0.0862), indicating the effectiveness of the proposed graph-driven cross-modal alignment mechanism.

Overall, these results demonstrate that SGR-BIM delivers consistently higher performance across all stages and question categories, with particularly strong advantages in tasks that require multi-hop rule interpretation and geometry-intensive reasoning. The finer-grained category-level analysis in Section 4.3.2 further examines how task complexity influences these performance patterns.

### 4.3.2. Fine-grained Category Performance

To further investigate how task structure and reasoning complexity affect model behavior, we conduct a fine-grained analysis within each question category. The detailed quantitative results are presented across four tables: rule queries (Table 6), rule requirement queries (Table 7), model queries (Table 8), and check queries (Table 9). These tables collectively illustrate how accuracy varies with increasing logical complexity, geometric dependency, and the degree of cross-modal alignment required. Across all four subtasks, SGR-BIM consistently demonstrates high robustness, particularly in conditions involving multi-hop rule interpretation, geometric computation, or IFC entity relationship reasoning.

Table 6 Performance on RULE queries under different numbers of regulatory clause sources.

| Number of Rule Sources | Baseline | Number | Correct(%) | Partial(%) | Wrong(%) | Avg Score |
|---|---|---|---|---|---|---|
| 1 | Basic-tool agent | 23 | 87.0 | 4.3 | 8.7 | **89.1** |
| | Enhanced-tool agent | | 95.7 | 4.3 | 0.0 | **97.8** |
| | Camel Baseline | | 95.7 | 0.0 | 4.3 | **95.7** |
| | metaGPT Baseline | | 73.9 | 21.7 | 4.3 | **84.8** |
| | Autogen Baseline | | 91.3 | 4.3 | 4.3 | **93.5** |
| | **SGR-BIM** | | **95.7** | **0.0** | **4.3** | 95.7 |
| 2 | Basic-tool agent | 70 | 67.1 | 18.6 | 14.3 | **76.4** |
| | Enhanced-tool agent | | 78.6 | 8.6 | 12.8 | **82.8** |
| | Camel Baseline | | 78.6 | 5.7 | 15.7 | **81.4** |
| | metaGPT Baseline | | 70.0 | 15.7 | 14.3 | **77.9** |
| | Autogen Baseline | | 71.4 | 4.3 | 24.3 | **73.6** |
| | **SGR-BIM** | | **85.7** | **8.6** | **5.7** | **90.0** |
| 3 | Basic-tool agent | 20 | 65.0 | 15.0 | 20.0 | **72.5** |
| | Enhanced-tool agent | | 80.0 | 15.0 | 5.0 | **87.5** |
| | Camel Baseline | | 75.0 | 10.0 | 15.0 | **80.0** |
| | metaGPT Baseline | | 85.0 | 5.0 | 10.0 | **87.5** |
| | Autogen Baseline | | 80.0 | 10.0 | 10.0 | **85.0** |
| | **SGR-BIM** | | **85.0** | **5.0** | **10.0** | **87.5** |
| **Overall** | Basic-tool agent | 113 | 70.8 | 15.0 | 14.2 | **78.3** |
| | Enhanced-tool agent | | 82.3 | 8.8 | 8.8 | **86.7** |
| | Camel Baseline | | 81.4 | 5.3 | 13.3 | **84.1** |
| | metaGPT Baseline | | 73.5 | 15.0 | 11.5 | **81.0** |
| | Autogen Baseline | | 77.0 | 5.3 | 17.7 | **79.6** |
| | **SGR-BIM** | | **87.6** | **6.2** | **6.2** | **90.7** |

The results for rule queries, summarized in Table 6, reveal how performance changes as the number of required rule sources increases. When only a single regulatory clause is needed, all systems perform relatively well, with most models, including CAMEL, AutoGen, and Enhanced-tool Agent, scoring above 93%, while SGR-BIM achieves 95.7%. However, when two clauses are required, accuracy declines noticeably for all baselines, while SGR-BIM remains substantially more stable (90.0%). This stability persists in the most challenging three-clause setting, where SGR-BIM reaches 87.5%, whereas the best-performing multi-agent baseline (CAMEL) falls to 80.0%, and MetaGPT drops to 68.4%.

Table 7 Performance on RULE REQUIREMENT queries across increasing geometric and relational reasoning complexity.

| **Calculation or Element Relationship Involved** | **Baseline** | **Number** | **Correct(%)** | **Partial(%)** | **Wrong(%)** | **Avg Score** |
|---|---|---|---|---|---|---|
| No calculation and relationship | Basic-tool agent | 87 | 0.0 | 11.5 | 88.5 | **5.7** |
| | Enhanced-tool agent | | 57.5 | 10.3 | 32.2 | **62.6** |
| | Camel Baseline | | 57.5 | 10.3 | 32.2 | **62.6** |
| | metaGPT Baseline | | 13.8 | 59.8 | 26.4 | **43.7** |
| | Autogen Baseline | | 42.5 | 19.5 | 37.9 | **52.3** |
| | **SGR-BIM** | | **65.5** | **16.1** | **18.4** | **73.6** |
| Calculation | Basic-tool agent | 83 | 0.0 | 21.7 | 78.3 | **10.8** |
| | Enhanced-tool agent | | 68.7 | 9.6 | 21.7 | **73.5** |
| | Camel Baseline | | 59.0 | 13.3 | 27.7 | **65.7** |
| | metaGPT Baseline | | 34.9 | 36.1 | 28.9 | **53.0** |
| | Autogen Baseline | | 50.6 | 21.7 | 27.7 | **61.4** |
| | **SGR-BIM** | | **74.7** | **12.0** | **13.3** | **80.7** |
| Relationship and Calculation | Basic-tool agent | 19 | 0.0 | 15.8 | 84.2 | **7.9** |
| | Enhanced-tool agent | | 89.5 | 5.3 | 5.3 | **92.1** |
| | Camel Baseline | | 84.2 | 0.0 | 15.8 | **84.2** |
| | metaGPT Baseline | | 57.9 | 21.1 | 21.1 | **68.4** |
| | Autogen Baseline | | 84.2 | 5.3 | 10.5 | **86.8** |
| | **SGR-BIM** | | **89.5** | **5.3** | **5.3** | **92.1** |
| **Overall** | Basic-tool agent | 189 | 0.0 | 16.4 | 83.6 | **8.2** |
| | Enhanced-tool agent | | 65.6 | 9.5 | 24.9 | **70.4** |
| | Camel Baseline | | 60.8 | 10.6 | 28.6 | **66.1** |
| | metaGPT Baseline | | 27.5 | 45.5 | 27.0 | **50.3** |
| | Autogen Baseline | | 50.3 | 19.0 | 30.7 | **59.8** |
| | **SGR-BIM** | | **72.0** | **13.2** | **14.8** | **78.6** |

A similar pattern is observed for rule requirement queries, detailed in Table 7, which additionally require inferring the geometric specifications needed for design compliance. In the simplest subset, queries requiring neither calculation nor cross-entity reasoning, SGR-BIM achieves 73.6%, outperforming CAMEL (62.6%) and AutoGen (52.3%). When geometric calculation is required, the gap widens significantly, with SGR-BIM achieving 80.7% compared to CAMEL (65.7%), MetaGPT (53.0%), and Autogen (61.4%). The most challenging queries, involving both calculations and spatial relationship inference, further highlight SGR-BIM's advantage: it reaches 92.1%, matching or exceeding the strongest baseline conditions and significantly outperforming MetaGPT (68.4%). Overall, SGR-BIM attains 78.6% on this category, outperforming all baseline models.

Table 8 Performance on MODEL queries across different geometric computation and relational reasoning conditions.

| Calculation or Element Relationship Involved | Baseline | Number | Correct(%) | Partial(%) | Wrong(%) | Avg Score |
|---|---|---|---|---|---|---|
| No calculation and relationship | Basic-tool agent | 55 | 0.0 | 10.9 | 89.1 | **5.5** |
| | Enhanced-tool agent | | 74.5 | 5.5 | 20.0 | **77.3** |
| | Camel Baseline | | 72.7 | 0.0 | 27.3 | **72.7** |
| | metaGPT Baseline | | 74.5 | 9.1 | 16.4 | **79.1** |
| | Autogen Baseline | | 58.2 | 27.3 | 14.5 | **71.8** |
| | **SGR-BIM** | | **85.5** | **3.6** | **10.9** | **87.3** |
| Calculation | Basic-tool agent | 89 | 13.5 | 4.5 | 82.0 | **15.7** |
| | Enhanced-tool agent | | 75.3 | 12.4 | 12.4 | **81.5** |
| | Camel Baseline | | 82.0 | 4.5 | 13.5 | **84.3** |
| | metaGPT Baseline | | 64.0 | 2.2 | 33.7 | **65.2** |
| | Autogen Baseline | | 82.0 | 3.4 | 14.6 | **83.7** |
| | **SGR-BIM** | | **88.8** | **6.7** | **4.5** | **92.1** |
| Relationship | Basic-tool agent | 46 | 2.2 | 6.5 | 91.3 | **5.4** |
| | Enhanced-tool agent | | 71.7 | 13.0 | 15.2 | **78.3** |
| | Camel Baseline | | 76.1 | 0.0 | 23.9 | **76.1** |
| | metaGPT Baseline | | 76.1 | 4.3 | 19.6 | **78.3** |
| | Autogen Baseline | | 76.1 | 4.3 | 19.6 | **78.3** |
| | **SGR-BIM** | | **89.1** | **2.2** | **8.7** | **90.2** |
| Relationship and Calculation | Basic-tool agent | 50 | 0.0 | 2.0 | 98.0 | **1.0** |
| | Enhanced-tool agent | | 56.0 | 18.0 | 26.0 | **65.0** |
| | Camel Baseline | | 70.0 | 0.0 | 30.0 | **70.0** |
| | metaGPT Baseline | | 68.0 | 6.0 | 26.0 | **71.0** |
| | Autogen Baseline | | 80.0 | 4.0 | 16.0 | **82.0** |
| | **SGR-BIM** | | **78.0** | **2.0** | **20.0** | **79.0** |
| **Overall** | Basic-tool agent | 240 | 5.4 | 5.8 | 88.8 | **8.3** |
| | Enhanced-tool agent | | 70.4 | 12.1 | 17.5 | **76.5** |
| | Camel Baseline | | 76.2 | 1.7 | 22.1 | **77.1** |
| | metaGPT Baseline | | 69.6 | 5.0 | 25.4 | **72.1** |
| | Autogen Baseline | | 75.0 | 9.2 | 15.8 | **79.6** |
| | **SGR-BIM** | | **85.8** | **4.2** | **10.0** | **87.9** |

For model queries, summarized in Table 8, which focus on extracting geometric or relational information from the IFC model, SGR-BIM again shows the best performance. In tasks with no calculation or relationship reasoning, SGR-BIM achieves 87.3%, a clear improvement over CAMEL (72.7%) and MetaGPT (79.1%). When calculation is required, SGR-BIM reaches 92.1%, surpassing CAMEL (84.3%) and AutoGen (83.7%), with MetaGPT dropping significantly to 65.2%. Relationship-only tasks show a similar advantage, with SGR-BIM attaining 90.2%, compared with typical multi-agent baseline scores ranging from 76–78%. Even in the most complex combined relationship-and-calculation subset, SGR-BIM maintains competitive performance at 79.0%, while Enhanced-tool Agent drops to 65.0% and MetaGPT to 71.0%.

Aggregated across all model queries, SGR-BIM achieves 87.9%, outperforming all baselines in model-centric queries.

Table 9 Performance on CHECK queries across different geometric computation and relational reasoning conditions.

| **Calculation or Element Relationship Involved** | **Baseline** | **Number** | **Correct(%)** | **Partial(%)** | **Wrong(%)** | **Avg Score** |
|---|---|---|---|---|---|---|
| No calculation and relationship | Basic-tool agent | 20 | 0.0 | 0.0 | 100.0 | **0.0** |
| | Enhanced-tool agent | | 35.0 | 10.0 | 55.0 | **40.0** |
| | Camel Baseline | | 40.0 | 35.0 | 25.0 | **57.5** |
| | metaGPT Baseline | | 25.0 | 25.0 | 50.0 | **37.5** |
| | Autogen Baseline | | 25.0 | 15.0 | 60.0 | **32.5** |
| | **SGR-BIM** | | **80.0** | **0.0** | **20.0** | **80.0** |
| Calculation | Basic-tool agent | 56 | 0.0 | 5.4 | 94.6 | **2.7** |
| | Enhanced-tool agent | | 75.0 | 12.5 | 12.5 | **81.2** |
| | Camel Baseline | | 60.7 | 14.3 | 25.0 | **67.9** |
| | metaGPT Baseline | | 44.6 | 32.1 | 23.2 | **60.7** |
| | Autogen Baseline | | 48.2 | 5.4 | 46.4 | **50.9** |
| | **SGR-BIM** | | **71.4** | **16.1** | **12.5** | **79.5** |
| Relationship and Calculation | Basic-tool agent | 61 | 0.0 | 13.1 | 86.9 | **6.6** |
| | Enhanced-tool agent | | 68.9 | 13.1 | 18.0 | **75.4** |
| | Camel Baseline | | 60.7 | 18.0 | 21.3 | **69.7** |
| | metaGPT Baseline | | 59.0 | 13.1 | 27.9 | **65.6** |
| | Autogen Baseline | | 77.0 | 1.6 | 21.3 | **77.9** |
| | **SGR-BIM** | | **77.0** | **9.8** | **13.1** | **82.0** |
| **Overall** | Basic-tool agent | 137 | 0.0 | 8.0 | 92.0 | **4.0** |
| | Enhanced-tool agent | | 66.4 | 12.4 | 21.2 | **72.6** |
| | Camel Baseline | | 57.7 | 19.0 | 23.4 | **67.2** |
| | metaGPT Baseline | | 48.2 | 22.6 | 29.2 | **59.5** |
| | Autogen Baseline | | 57.7 | 5.1 | 37.2 | **60.2** |
| | **SGR-BIM** | | **75.2** | **10.9** | **13.9** | **80.7** |

Finally, the most challenging category, check queries, is presented in Table 9. These tasks integrate rule interpretation, model extraction, geometric computation, and cross-entity relationship reasoning. In the simplest subset requiring neither calculation nor relationship reasoning, baseline accuracy varies widely: the Basic-tool Agent performs poorly (0%), CAMEL moderately (57.5%), and AutoGen even lower (32.5%). In contrast, SGR-BIM achieves 80.0%, demonstrating improved robustness even for simpler compliance judgments. For calculation-based compliance checks, SGR-BIM again leads with 79.5%, while CAMEL reaches 67.9% and AutoGen only 50.9%. The most demanding relationship-and-calculation subset further highlights the strength of explicit alignment: SGR-BIM reaches 82.0%, outperforming CAMEL (69.7%) and MetaGPT (65.6%). Overall, SGR-BIM achieves 80.7% in this category, exceeding all baselines.

Across all four tables and all fine-grained difficulty subdivisions, the results show a consistent pattern: SGR-BIM remains substantially more stable than all multi-agent baselines as task complexity increases, particularly when reasoning requires multi-clause regulation interpretation, geometric computation, or latent relationship resolution across IFC entities. These findings reinforce the overall performance trends reported in Section 4.3.1 and illustrate the critical role of explicit graph-driven semantic alignment in enabling reliable geometry compliance check.

### 4.4. Error Analysis

To better understand the limitations of the proposed framework, we conducted a detailed error analysis on all responses whose accuracy label was not equal to 1 (i.e., responses assessed as partially correct with a score of 0.5 or incorrect with a score of 0). Across the full dataset of 679 queries, 135 responses fell into this category, corresponding to an overall error proportion of 19.9%. These cases were further categorized into five distinct error types: incomplete answers, misunderstanding of user intent, insufficient understanding of model or rule information, faulty reasoning and inference, and systemic errors related to tool or system behavior. The distribution of these error types across all task categories is summarized in Table 10(a). As illustrated in Table 10(b), a fully correct score (1) is strictly reserved for responses that not only reach the correct conclusion but also demonstrate a complete mastery of the underlying regulatory logic, such as identifying the relationship between area, occupancy factors, and discrete capacity thresholds.

Table 10(a) Distribution of error types across different question categories.

| Error Category/Category | Rule | Rule Requirement | Model | Check | Overall |
|---|---|---|---|---|---|
| Number | 14 | 53 | 34 | 34 | 135 |
| Total Number | 113 | 189 | 240 | 137 | 679 |
| Incomplete answer | 50.0% | 47.2% | 29.4% | 44.1% | 42.2% |
| Misunderstanding of user intent | 7.1% | 5.7% | 2.9% | 0% | 3.7% |
| Insufficient understanding of model or rule information | 0% | 22.6% | 17.6% | 5.9% | 14.8% |
| Faulty reasoning and inference | 42.9% | 18.9% | 14.7% | 35.3% | 24.4% |
| Systemic error | 0% | 5.7% | 35.3% | 14.7% | 14.8% |

Table 10(b) Representative scoring examples for the Accuracy metric.

| Score | Rating | Response Example (Query: "Do larger rooms automatically require additional exit doors according to the code?") | Rationale for Score |
|---|---|---|---|
| 1 | Fully Correct | **No, larger rooms do not automatically require additional exit doors.** Exit door requirements are based on occupant capacity (room area / occupancy factor). A larger room with a high occupancy factor (e.g., a warehouse at 30 m²/person) may require fewer exit doors than a smaller room with a low occupancy factor (e.g., a restaurant at 1 m²/person). The trigger is the calculated occupant capacity, not room size alone. Also, even if the occupancy factor is constant, an increase in area only necessitates more doors if the resulting capacity crosses a specific threshold. | Conclusion and logic are both fully correct. The response demonstrates a multi-variable analysis: it explains the compensatory effect of the occupancy factor and identifies that door requirements are tied to discrete capacity thresholds. It correctly refutes the "automatic" premise by highlighting that area increases do not always lead to capacity-threshold crossings. |
| 0.5 | Partially Correct | **Yes, larger rooms do require additional exit doors.** The requirement is based on occupant capacity rather than room area alone. The number of exit doors increases at specific occupant capacity thresholds: 1 door for 4–30 persons, 2 doors for 31–500 persons, 3 | Correctly identifies the core mechanism (occupant capacity and thresholds). However, it fails to reach the precise conclusion regarding the "automatic" nature of the query. It |

| | | doors for 501–750 persons, and 4 doors for 751–1000 persons. | misses the comparative discussion on how different occupancy factors can result in larger rooms requiring fewer doors than smaller ones. |
|---|---|---|---|
| **0** | **Incorrect** | **Yes, larger rooms automatically require additional exit doors.** The requirement is based on calculated occupant capacity. The number of exit doors increases with higher occupant counts, which are calculated by dividing room area by the appropriate occupancy factor. | Incorrect conclusion and flawed logic. While it mentions capacity, it fundamentally confirms the "automatic" premise and fails to distinguish between a continuous area increase and the discrete, factor-dependent nature of the regulatory requirement. |

Table 10(c) Diagnostic breakdown of error patterns.

| Error Category | Query | Answer (Incorrect/Partial) | Ground Truth | Explanation of Error |
|---|---|---|---|---|
| **Incomplete Answer** | "Do larger rooms automatically require additional exit doors according to the code?" | "Yes, larger rooms do require additional exit doors. The requirement is based on occupant capacity... [lists thresholds]." | "No, larger rooms do not automatically require additional exits. It depends on the occupant capacity (Area/Factor)... a larger room with a high factor may require fewer doors than a smaller room." | The system identifies the capacity-based mechanism but fails to recognize the logic of thresholds and the comparative effect of occupancy factors, leading to an imprecise conclusion. |
| **Misunderstanding of User Intent** | "Does the area behind this door need to provide at least two exits, as required by the code?" | "Yes, the area behind this door (Living room) meets the exit requirements. The room has 1 door with a width of **1.81m**, exceeding the **750mm** requirement." | "No, the area only needs 1 exit door with a width of at least **750mm** for its 3-person occupant capacity." | The system misinterprets a query about the **required quantity** of exits as a compliance check for the **existing door's width**, failing to address the core question of whether a second exit is mandatory. |
| **Insufficient Information Retrieval** | "How wide does this exit door is required to be according to the code?" | "According to rules for warehouses, the factor is **30m²/person**. However, without specific area information, I cannot calculate the exact capacity." | "The door must be at least **750mm** wide. The room area is **80m²**, resulting in a capacity of 3 persons, which falls into the **750mm** requirement bracket." | The system identifies the correct regulatory logic but fails to invoke the tool necessary to retrieve specific BIM data (room area), resulting in a generic explanation instead of a specific calculation. |
| **Faulty Reasoning and Inference** | "Which fire compartments are on Floor 1?" | "Based on the room naming convention, I can see that rooms starting with '1-' are on Floor 1. | "There are 5 fire compartments on Floor 1 [lists names]. These were identified by querying the explicit | The system relies on a heuristic "naming convention" shortcut rather than |

| | | | |
|---|---|---|---|
| | Therefore, the fire compartments on Floor 1 are..." | 'contains' relationships in the IFC model." | executing a formal query of the IFC spatial structure, which is unreliable for non-standardized models. |

Across all categories, incomplete answers represent the most frequent error type, accounting for 42.2% of all erroneous responses. This issue is especially prevalent in rule requirement (47.2%) and rule queries (50.0%). These tasks typically involve multi-clause rule interpretation and extraction of design thresholds from textual or tabular specifications. Errors in this category most often arise when the system retrieves the correct rule fragments but fails to consolidate all relevant conditions, such as missing exceptions, scope restrictions, or numerical thresholds, leading to partially correct or underspecified outputs. The "Incomplete Answer" case in Table 10(c) further elucidates this phenomenon, where the system correctly identifies the capacity-based mechanism but fails to recognize the effect of occupancy factors, resulting in an over-generalized conclusion. Although SGR-BIM reduces such omissions compared with baselines, the persistence of this issue highlights the inherent difficulty of aggregating dispersed regulatory information into a complete and coherent answer.

Faulty reasoning and inference, the second largest error category (24.4%), is particularly frequent in rule (42.9%) and check tasks (35.3%). These tasks require multi-hop logical chaining that spans rule semantics, geometric attributes, and IFC entity relationships. Typical mistakes include applying incorrect occupancy factors or using incorrect attributes from certain elements. As diagnosed in Table 10(c), these errors often stem from the model's reliance on unreliable heuristics, such as room naming conventions, rather than executing formal spatial queries within the IFC structure. This indicates that while graph alignment provides structured context, robust topological navigation remains a challenge for agents.

A third category, insufficient understanding of model or rule information, accounts for 14.8% of errors, especially in rule requirement (22.6%) and model queries (17.6%). Errors here often involve referencing the wrong IFC entity, retrieving an incorrect geometric attribute, or overlooking a key parameter in regulatory tables. A representative instance is provided in Table 10(c), where the system correctly interprets the regulatory logic but fails to invoke the necessary tool to retrieve the required room area from the BIM model, leading to an abstract response rather than a grounded calculation. This pattern suggests that although cross-modal alignment provides explicit connections between rules and geometry, fine-grained data extraction still presents difficulties, particularly when IFC entities have heterogeneous modeling patterns or when multiple candidate attributes appear relevant.

Misunderstanding of user intent, though relatively infrequent (3.7%), occurs across three categories. These errors typically arise when the system misclassifies a query, such as interpreting a requirement query as a compliance check. Table 10(c) highlights such a "semantic drift," where the system focuses on checking a door's width instead of addressing the user's core question regarding the mandatory quantity of exits. While the occurrence rate is low, such misinterpretations can lead to fully irrelevant answers, indicating that accurate intent grounding remains essential for reliable interactive behavior.

Finally, systemic errors (14.8% overall) reflect issues not with reasoning but with system execution, such as failures in IFC parsing, incorrect tool responses, or inconsistencies in intermediate agent states. These errors are most prominent in model queries (35.3%), where the IFC analyzing toolkit interacts heavily with the BIM model. Although such issues lie outside the core reasoning mechanism, they emphasize the importance of robust engineering when deploying geometry compliance checking systems on real-world IFC data.

Overall, the error analysis indicates that the dominant sources of errors are incomplete synthesis of information and failures in complex multi-step reasoning, both central challenges of geometry compliance check for fire safety codes. Compared with these challenges, errors arising from misunderstanding user intent occur far less frequently, suggesting that the graph-driven alignment mechanism is generally effective at grounding queries and linking them to corresponding rule and model contexts. These findings highlight that the main opportunities for future improvement lie in strengthening the system's ability to support deeper multi-hop reasoning and ensure more reliable extraction and integration of BIM-based geometric information within complex compliance check tasks.

### 4.5. Discussion

To further assess the robustness and utility of the proposed framework, this section provides a discussion of the experimental results.

#### 4.5.1. Impact of Backend LLM Choice

While the main experiments in this study are conducted using DeepSeek-3.2 [59] as the backend language model, we additionally examine the impact of backend LLM choice using part of the samples from the whole dataset by evaluating SGR-BIM with GPT-4o [60]. This analysis serves as a complementary investigation of proposed framework between language models. The results are summarized in Table 11.

Table 11 Performance of SGR-BIM under different LLM backends.

| Category | Backend | Correct (%) | Partial (%) | Wrong (%) | Avg Score |
|---|---|---|---|---|---|
| **Rule** | GPT-4o | 83.2 | 8.0 | 8.8 | **87.2** |
| | DeepSeek-3.2 | 87.6 | 6.2 | 6.2 | **90.7** |
| **Rule Requirement** | GPT-4o | 46.5 | 20.9 | 32.6 | **57** |
| | DeepSeek-3.2 | 67.4 | 18.6 | 14.0 | **76.7** |
| **Model** | GPT-4o | 69.6 | 7.1 | 23.2 | **73.2** |
| | DeepSeek-3.2 | 87.5 | 7.1 | 5.4 | **91.1** |
| **Check** | GPT-4o | 43.3 | 16.7 | 40.0 | **51.7** |
| | DeepSeek-3.2 | 80.0 | 10.0 | 10.0 | **85.0** |
| **Overall** | GPT-4o | 68.6 | 11.2 | 20.2 | **74.2** |
| | DeepSeek-3.2 | 83.1 | 9.1 | 7.9 | **87.6** |

Overall, SGR-BIM exhibits consistent performance trends across both backends, with DeepSeek-3.2 yielding higher accuracy across all categories. The performance difference is relatively small for rule queries, where both backends achieve strong results once regulatory semantics are explicitly structured by the graph-based alignment mechanism. This observation suggests that for tasks dominated by rule retrieval and interpretation, the framework's alignment strategy plays a more decisive role than the specific language model used.

In contrast, larger performance gaps emerge in categories that require deeper geometric reasoning and cross-modal integration, including rule requirement, model, and geometry compliance check tasks. In these settings, DeepSeek-3.2 consistently achieves higher accuracy and lower error rates than GPT-4o. This pattern indicates that backend models differ in their ability to sustain multi-step reasoning over structured geometric information and to reliably integrate intermediate results across multiple reasoning steps. From a methodological perspective, these results suggest that differences in backend performance mainly affect the absolute accuracy level rather than the overall behavior or strengths of the framework.

#### 4.5.2. Evaluation on Explainability and Reasoning Quality

Beyond accuracy, practical geometry compliance check requires that system outputs be coherent, relevant, and explainable, particularly in regulatory and safety-critical contexts where users must understand not only what decision is made, but also why. To examine these qualitative aspects, we evaluate SGR-BIM using additional metrics, coherence, relevance, and explainability, across all task categories. Rather than comparing against multiple multi-agent frameworks, this analysis focuses exclusively on the Enhanced-tool Agent as the baseline. This choice isolates the impact of multi-agent coordination and reasoning organization, since both systems have access to the same domain-specific toolset, and differ primarily in how retrieved information is aligned, integrated, and reasoned over.

Table 12 Qualitative evaluation of explainability and reasoning quality.

| Category | Agent | Accuracy | Coherence | Relevance | Explainability |
|---|---|---|---|---|---|
| Rule | Enhanced-tool Agent | 86.7 | 95.9 | 98.8 | 65.5 |
| | SGR-BIM | 90.7 | 96.8 | 97.1 | 97.3 |
| Rule Requirement | Enhanced-tool Agent | 70.4 | 71.4 | 92.5 | 71.8 |
| | SGR-BIM | 78.6 | 96.1 | 96.8 | 99.3 |
| Model | Enhanced-tool Agent | 76.5 | 74.9 | 97.3 | 55.8 |
| | SGR-BIM | 87.9 | 87.9 | 99.4 | 99.9 |
| Check | Enhanced-tool Agent | 72.6 | 83 | 99.2 | 97.5 |

| | SGR-BIM | 80.7 | 97.7 | 100 | 100 |
|---|---|---|---|---|---|
| **Overall** | Enhanced-tool Agent | 75.7 | 79.2 | 96.7 | 70.4 |
| | SGR-BIM | 84.3 | 93.7 | 98.4 | 99.3 |

These metrics except accuracy are derived through LLM-assisted evaluation following predefined criteria and subsequently verified through manual inspection to ensure consistency and correctness. This inspection was conducted by a research team consisting of three members with BIM expertise, all of whom possess over one year of experience in BIM research, including one senior practitioner with extensive industry experience. To ensure the objectivity of the manual audit, the reviewers first performed independent checks on the LLM-generated scores. Any discrepancies or borderline cases between the LLM judge and human reviewers, or among the human reviewers themselves, were resolved through a formal consensus-based reconciliation process. In these sessions, the team collectively reviewed the reasoning path and the IFC model context to determine the final score.

This evaluation setup allows us to assess qualitative reasoning characteristics while maintaining practical scalability. The results, summarized in Table 12, show that SGR-BIM consistently outperforms the Enhanced-tool Agent across all non-accuracy metrics and all task categories. For coherence, which measures whether answers fully and fluently cover the necessary aspects of a question, SGR-BIM achieves substantial gains. This improvement is especially evident in rule requirement and check categories, where regulatory conditions, geometric constraints, and contextual qualifiers must be combined into a unified response. The Enhanced-tool Agent, despite accessing correct information, often produces fragmented or partially articulated answers, whereas SGR-BIM benefits from graph-based aggregation of relevant elements, leading to more complete and structured outputs. Performance differences are less pronounced for relevance, where both systems perform strongly. This is expected, as both models leverage explicit tool-based retrieval, ensuring that answers generally address the intended question.

The most striking contrast appears in explainability. Across all categories, SGR-BIM achieves high explainability scores, while the Enhanced-tool Agent exhibits significantly lower values, particularly in model and rule queries. In many cases, the Enhanced-tool Agent provides correct conclusions but fails to articulate the intermediate reasoning steps, leaving key assumptions, calculations, or rule references implicit. In contrast, SGR-BIM generates explicit reasoning traces grounded in the semantic graph, making intermediate decisions, referenced entities, and applied constraints transparent and easy to follow. This difference directly reflects the effect of graph-driven alignment and multi-agent coordination, which encourage the system to reason over structured representations rather than relying solely on implicit chain-of-thought generation.

Importantly, this explainability is not limited to textual output alone. The structured reasoning paths generated over the semantic graph are directly reflected in the interactive user interface introduced earlier. Through the UI, users can inspect referenced BIM elements, trace which regulatory clauses were applied, and review intermediate geometric evaluations, thereby transforming explainability from a passive narrative into an interactive exploration process. This tight coupling between graph-based reasoning and user-facing visualization enables practitioners to verify results, identify potential modeling issues, and iteratively refine designs with greater confidence.

Overall, this evaluation demonstrates that the benefits of SGR-BIM extend beyond improved accuracy. By explicitly structuring cross-modal information and organizing reasoning over a shared semantic graph, the framework substantially enhances answer coherence and explainability without sacrificing relevance. These qualities are essential for real-world fire safety compliance workflows, where regulatory approval, design iteration, and professional accountability depend on transparent and interpretable decision-making. The results further support the argument that structured alignment is a key enabler of explainable and trustworthy geometry compliance check.

#### 4.5.3. Analysis of Knowledge Graph Characteristics

To provide a clearer understanding of the behavior and scalability of the knowledge graph, this section presents an analysis of its structural characteristics across different query categories. In SGR-BIM, the knowledge graph is implemented as a dynamic directed multi-graph, whose size and topology evolve iteratively depending on the query intent, regulatory complexity, and BIM model structure. Rather than constructing a full-scale representation of the BIM model, the graph is built in a task-driven manner, incrementally incorporating only the entities and relationships required for the current compliance evaluation.

Table 13 Knowledge Graph Statistics across Different Query Categories

| Stage | Category | Avg. Nodes* | Avg. Edges* | Avg. Hop Distance* | Typical Range |
|---|---|---|---|---|---|
| Design-stage | Rule | ~ 3 - 5 | ~ 2 - 4 | ~ 3 - 5 | Small |
| | Rule Requirement | ~ 5 - 7 | ~ 5 - 7 | ~ 5 - 7 | Medium |
| Check-stage | Model | ~ 5 - 7 | ~ 5 - 7 | ~ 5 - 7 | Medium |
| | Check | ≥ 7 | ≥ 7 | ≥ 7 | Large |

*Note: The reported values represent approximate ranges and may vary depending on query complexity and BIM model size. For example, when evaluating a stair serving multiple rooms or fire compartments in complex BIM models, the number of associated entities can increase substantially, resulting in rapid growth in node and edge counts.

Table 13 summarizes the observed graph statistics across different query categories defined in Section 4.1. The results show that the graph size varies systematically with the category of the task. For simple rule-based queries (Rule category), the graph typically contains a small number of nodes and edges, as only regulatory information is required. In contrast, Rule Requirement and Model queries involve moderate graph sizes, reflecting the need to interpret constraints or retrieve model-specific attributes. The most complex category, Check queries, requires integrating regulatory provisions with BIM entities through multi-hop reasoning chains, resulting in larger graphs with increased numbers of nodes, edges, and reasoning steps. The reported hop distance corresponds to the length of the reasoning path connecting the initial query to the final decision, reflecting the depth of cross-modal dependencies involved in the task. It should be noted that the statistics reported in Table 13 represent approximate ranges observed during the experimental evaluation.

The size and structure of the graph are influenced by several factors. First, the query category plays a dominant role: queries that involve only a single information source (either regulatory codes or BIM model) tend to produce smaller graphs, while compliance checking tasks that require cross-modal alignment result in larger and more complex structures. Second, the complexity of the reasoning process directly affects graph expansion. Multi-step rule chains introduce intermediate entities and relationships, increasing both node and edge counts. Third, the scale of the BIM model also impacts graph size. For example, when evaluating relationships such as a stair serving multiple rooms or fire compartments, a large number of connected entities may be introduced into the graph, leading to significant growth in its structure.

The architecture of the knowledge graph is illustrated in Fig 4, which provides both the graph schema and a concrete example of its application in a check task. This figure demonstrates how nodes (e.g., BIM entities and regulatory rules) and relational edges are dynamically organized to support multi-hop reasoning. The query, "Does the width of this door comply with the code requirements?", serves as the root node from which the graph is incrementally constructed through iterative agent interactions. The graph is not generated in a single step. Instead, it is progressively expanded as the system retrieves, computes, and integrates information across modalities during the reasoning process.

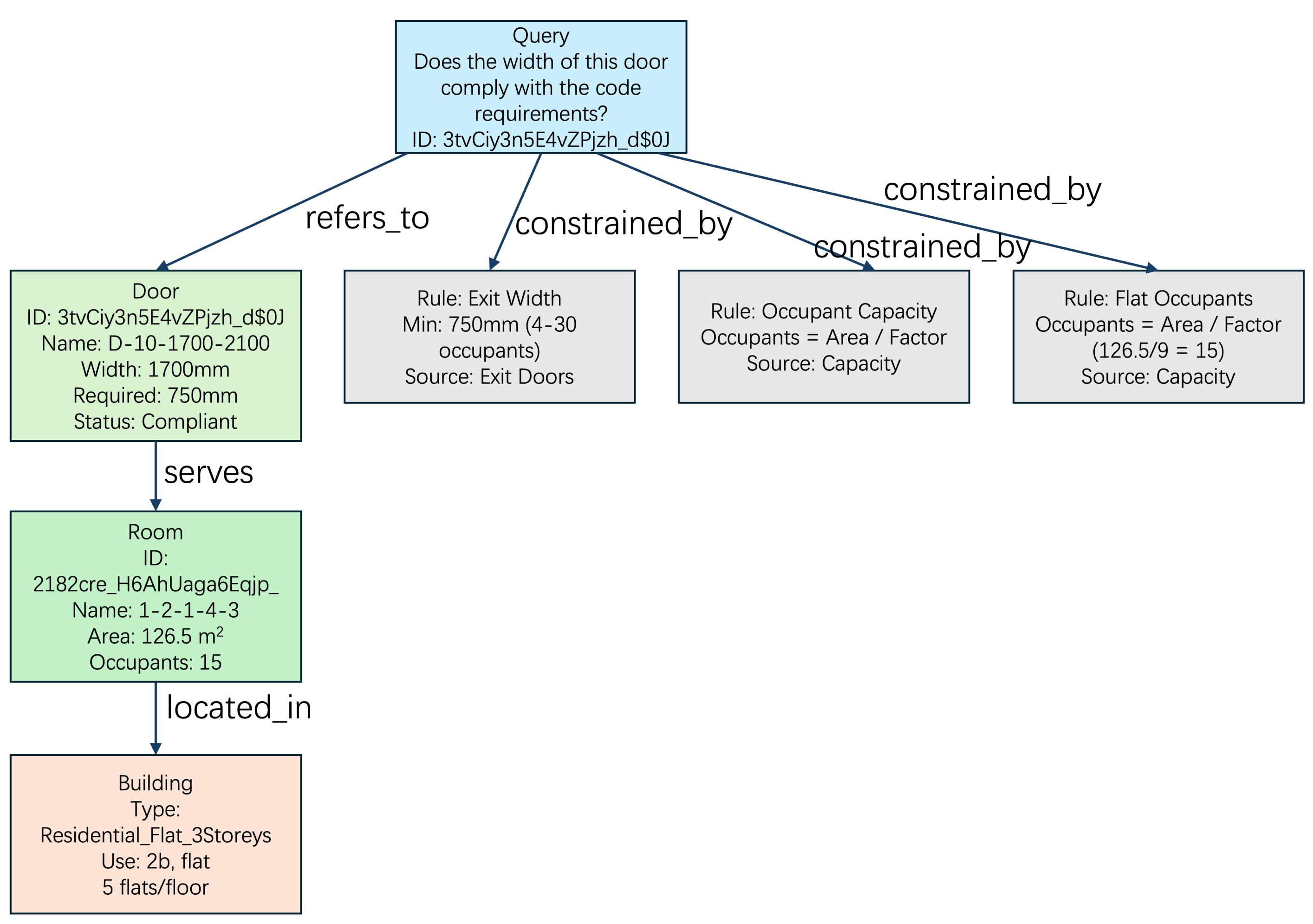

*Fig. 4 Structural schema and representative instance of the knowledge graph for a compliance query.*

Starting from the query node, the system first identifies the relevant BIM entity (Door) and retrieves its geometric attribute (width). However, the required width cannot be directly determined at this stage, as it depends on the occupant load associated with the connected space. This dependency triggers further graph expansion: the Door node is linked to its corresponding Room node, from which the area is obtained. The occupant capacity is then computed based on regulatory rules (e.g., occupancy factor), and only after this calculation is completed is the occupant attribute added to the Room node. This intermediate result subsequently enables the evaluation of the minimum required door width.

This incremental construction process ensures that all intermediate states, including derived attributes and rule dependencies, are explicitly preserved. Consequently, the graph serves not only as a data structure for reasoning but also as a transparent trace of the decision-making process, enabling users to inspect how each piece of information contributes to the final compliance result.

Finally, on the other hand, it is important to clarify the scope of geometric representation supported by the proposed method. In this study, geometric properties are derived from semantic and parametric information available in IFC models, such as dimensions, areas, and spatial relationships associated with building elements. The framework does not directly operate on low-level geometric representations such as tessellated meshes. If the IFC model encodes geometry purely as mesh data without explicit parametric attributes, additional pre-processing would be required to extract meaningful geometric information. This limitation defines the current applicability of the method and highlights a direction for future research in integrating mesh-level geometric reasoning.

**4.5.4. Functional Comparison and Practical Applicability Analysis**

To evaluate the practical utility of SGR-BIM within the broader AEC industry, we conducted a functional comparison of its system characteristics against three established Automated Compliance Checking (ACC) paradigms: commercial model checkers (e.g., Solibri), hard-coded/visual programming scripts (e.g., C# or Dynamo), and template-based research systems. This comparison, summarized in Table 14, clarifies the methodological shift introduced by SGR-BIM and highlights its unique advantages in real-world deployment scenarios.

Table 14 Functional Comparison and Practical Applicability Analysis

| **Comparison Dimensions** | **Commercial Model Checkers (e.g., Solibri)** | **Hard-coded/Visual Programming (e.g., C#, Dynamo)** | **Template-based Systems** | **SGR-BIM** |
|---|---|---|---|---|
| **User Interaction** | GUI-based Menu | Scripting Interface | Natural Language | Hybrid Conversational |

| | | | | and Visual Interface |
|---|---|---|---|---|
| **Rule Authoring Effort** | Low (if pre-defined) | High (Manual Coding) | Medium (Template Mapping) | Low (Agent-interpreted) |
| **Semantic Alignment** | Manual Mapping | Hard-coded Logic | Manual Mapping | Dynamic Graph-driven Alignment |
| **Geometric Flexibility** | Rigid (Fixed Operators) | High (Custom Algorithms) | Limited (Pre-defined Templates) | High (Autonomous Tool Use) |
| **Deployment Overhead** | High (Proprietary Setup) | High (Environment Config) | Medium | Low (API-integrated) |
| **Explainability and Transparency** | Black-box (Proprietary) | Code-level (Complex) | Fixed/Template-constrained | Traceable |

As shown in Table 14, commercial software and hard-coded systems rely on rigid, pre-defined operators or static semantic mapping, requiring users to possess high proficiency in either proprietary GUI configurations or domain-specific programming. SGR-BIM addresses this by shifting the interpretive burden from the user to autonomous agents. Through an hybrid conversational and visual interface, practitioners can engage with the compliance process using natural language while monitoring the agents' reasoning. This dynamic, graph-driven alignment mechanism allows for the resolution of complex, multi-hop regulatory queries without the need for manual rule-encoding or extensive template preparation.

Furthermore, the practical applicability of SGR-BIM is significantly enhanced by its focus on transparency. While proprietary commercial engines often operate as black boxes and complex scripts are difficult to audit, SGR-BIM provides traceable reasoning paths. By documenting the logical alignment between specific regulatory clauses and IFC entities in a human-readable format, the framework enables human-in-the-loop verification. This ensures that automated decisions are verifiable by domain experts, bridging the trust gap that often hinders the adoption of AI-driven tools in professional workflows. From a deployment perspective, the API-integrated nature of SGR-BIM offers a more scalable and flexible trajectory for integrating advanced reasoning capabilities into existing BIM environments, representing an advancement over rigid, siloed traditional engines.

## 5. Limitations and Future Work

Despite the performance improvements, this study has several limitations that provide directions for future research. First, the comparative analysis reveals that an enhanced-tool single agent already handles over 75% of queries correctly, suggesting that the current multi-agent graph alignment provides diminishing returns for simpler tasks. Future work should investigate more efficient adaptive architectures that only trigger complex multi-agent coordination when high reasoning depth is detected.

Second, the error analysis in Section 4.4 reveals that incomplete synthesis (42.2%) and faulty reasoning (24.4%) remain the primary failure modes. These discrepancies indicate that while the current graph-driven approach is effective, the system still encounters challenges in aggregating highly dispersed regulatory clauses and maintaining logical consistency over extended reasoning sequences. To mitigate these issues, future research will prioritize knowledge graph refinement, focusing on increasing the semantic granularity of nodes to better capture sub-clause dependencies and implicit geometric constraints within the code. Additionally, we will explore the optimization of memory mechanisms, such as implementing structured reasoning scratchpads or persistent context buffers for the agents. These enhancements aim to ensure that intermediate reasoning states are accurately preserved and verified across multi-hop traversals, thereby reducing logical drift in complex compliance tasks.

Third, the computational scalability of SGR-BIM to massive, high-density IFC models (e.g., high-rise healthcare facilities or complex industrial plants) requires further investigation. As the number of building entities and spatial relationships increases, the search space for graph alignment and the memory overhead for agentic coordination grow non-linearly. Subsequent studies will focus on developing hierarchical graph partitioning and spatial-semantic pre-indexing strategies to prune irrelevant sub-graphs and optimize traversal efficiency, ensuring the framework maintains high performance even as the building complexity scales to an industrial level.

In summary, while SGR-BIM establishes a transparent and graph-driven paradigm for automated compliance check, bridging the gap to full industrial deployment requires an iterative refinement of reasoning robustness and system efficiency. By systematically addressing the identified limitations, we aim to transition the framework from a research prototype into a resilient, industry-grade engine for trustworthy compliance check system in the AEC industry.

## 6. Conclusion

This paper addresses the fundamental challenge of semantic interoperability in BIM-based automated compliance checking (ACC), specifically for geometry-intensive regulations that rely on complex spatial contexts. To overcome the limitations of rigid, template-based methods, this paper presented SGR-BIM (Spatial-Geometric Reasoning System for BIM), which utilizes a dynamic cross-modal knowledge graph to autonomously align user intent with regulatory provisions and BIM geometry. This architecture supports transparent multi-hop reasoning and geometry-aware evaluation, transforming the checking process into an interpretable, interactive workflow. Validation on 679 queries demonstrates that SGR-BIM achieves 84.3% accuracy, representing an 8.6% improvement over enhanced-tool single-agent baselines. These findings underscore the value of graph-driven semantic alignment in handling intricate spatial constraints and provide an explainable foundation for digital design validation.

From a broader perspective, these results indicate that the technical bottleneck in ACC is transitioning from basic data extraction to the complex orchestration of geometric and semantic logic. Practically, the SGR-BIM framework provides a foundation for more interactive compliance tools, where architects can receive transparent reasoning paths rather than binary pass/fail results. This capability is particularly relevant for the early design stages, where real-time feedback on spatial constraints can significantly reduce the need for late-stage revisions and manual re-entry of occupant data.

However, several limitations warrant further research. Currently, the multi-agent coordination provides diminishing returns for simpler tasks, suggesting that future iterations should incorporate adaptive architectures to optimize computational efficiency. Moreover, to address logical consistency in highly dispersed regulatory sections, future work will focus on increasing the semantic granularity of knowledge graphs and implementing structured memory mechanisms. Finally, investigating hierarchical graph partitioning will be essential to ensure the scalability of this approach when applied to massive, high-density IFC models typical of industrial-scale projects.

**Acknowledgement**

The authors gratefully acknowledge the support provided by the Innovation and Technology Fund (Grant No. ITS/394/23FP) from the Innovation and Technology Support Program (ITSP) - Platform, Innovation and Technology Commission, HKSAR.

**Declaration of Generative AI and AI-assisted technologies in the writing process**

Statement: During the preparation of this work the author(s) used ChatGPT-5.1 in order to improve readability and language. After using this tool/service, the author(s) reviewed and edited the content as needed and take(s) full responsibility for the content of the publication.

**Data Availability Statement**

The data supporting the findings of this study are provided within the article and its Appendices. Specifically, Appendices A, B, and C provide comprehensive details regarding prompt design for LLM-based agents, knowledge graph implementation, and the evaluation protocol. These materials offer sufficient methodological transparency to facilitate a thorough understanding of the framework's reasoning logic and the verification of the experimental approach. The underlying source code and the complete dataset will be made available upon reasonable request.

## Appendix A. Implementation Details and Prompt Design for LLM-based Agents

This appendix provides detailed technical specifications of the LLM-based agents in the proposed SGR-BIM framework, including their implementation architecture, tool design, and prompt engineering strategies.

The SGR-BIM framework is implemented using the LangGraph library (https://github.com/langchain-ai/langgraph), which enables the construction of stateful, multi-agent workflows with explicit control over execution flow, memory, and iterative reasoning. The execution follows an iterative loop, where the Analysis Agent dynamically determines the next action (rule retrieval, IFC query, or termination), enabling adaptive multi-hop reasoning.

Each agent is implemented as a combination of a prompt template, which defines its role and reasoning behavior. For agents that require access to external information sources, additional tools are integrated to enable structured data retrieval (e.g., regulatory document search and IFC model querying). The prompt for all the agents is designed to follow four main principles: defining a clear role for a specific task, specifying the descriptions of the task, providing examples to complete the task, and constraining the output format to ensure consistency for downstream processing.

### A.1. User Intent Analysis Agent

The User Intent Analysis Agent is responsible for interpreting the user's natural language query and transforming it into a structured intent representation that can be processed within the semantic graph. Specifically, this agent identifies the target building element, the relevant compliance attributes, and any implicit constraints that are not explicitly stated in the query. This step establishes the initial query node in the graph and provides the foundation for subsequent cross-modal reasoning.

```
INTENT_AGENT_SYSTEM_PROMPT = """
You are a professional BIM query intent interpreter.
Your job is to convert the user's natural language question into a structured JSON object that describes the intent.
The JSON must contain exactly these fields:
1. "task_type": (string, required)
  - Allowed values:
```

• "query_model"  – Questions about the BIM model itself (no reference to rules).

• "query_rule"  – Questions about code/rule requirements in general (not tied to a specific element).

• "query_rule_requirement" – Questions asking what code/rule requirements apply to a specific element in the model.

• "check"  – Questions about whether a specific element complies with rules.

2. "target_entity": (string or list of strings, required)

- One or more of: "stair", "door", "room", "building", "floor", "seat", "bed", "fire compartment"

- Example: ["door","room"]

3. "attributes": (list of strings, required)

- One or more of: "area", "width", "total_width", "count", "class", "discharge_value", "occupant_capacity"

- Example: ["count","total_width"]

4. "context": (object or null, optional)

- Spatial or semantic qualifiers. May include:

• "floor": "2F"

• "building_type": "3a. Health/child care facilities"

• "room_id": "<GUID of an IfcSpace>"

• "zone_id": "<GUID of an IfcZone>"

- Example: { "floor":"2F"}

- If none apply, use null

- Do not invent missing details. If context is insufficient, simply set this field to null and leave further inference to downstream agents.

### ID Extraction and Typing (REQUIRED)

- Extract any explicit IDs (e.g., GUIDs) mentioned by the user.

- Determine the correct ID type based on the wording around the ID:

• Phrases like "this door", "Door D-102", or "door GUID ..." → use "door_id".

• Phrases like "this room/space", "Room A-101", or "space GUID ..." → use "room_id".

• Phrases like "this fire compartment/zone" → use "zone_id".

• Phrases like "this stair" → use "stair_id".

- If the question references one entity via another (e.g., "the room containing this door"), still assign the provided ID to its correct key (e.g., put the door GUID into "door_id"). Do not introduce intermediate relation hints; just type the provided ID correctly.

- If more than one ID is provided, include each under the appropriate key.

- If you cannot confidently determine the entity type of an ID, do not guess; set "context" to null.

### ID Field Conventions

- When the user refers to a room/space, always use **room_id** and pass the GUID of an **IfcSpace**.

- When the user refers to a fire compartment/zone, always use **zone_id** and pass the GUID of an **IfcZone**.

- Never pass a zone GUID where a room_id is expected, and never pass a room GUID where a zone_id is expected.

### Tool for Use Classification

When the user refers to a building type in plain language (for example, "6-storey flat" or "3-storey villa") and no code is provided, you should call:

`read_csv_table("Use Classification")`

to load the entire Use Classification table. Then inspect its rows to determine which Sub-category (e.g. "1b. Flats", "1a. House type dwellings") best matches the user's description, and set that code as:

`"context": { "building_type": "<matched Sub-category>" }`

Return only valid JSON (no markdown, no explanation).

Do not add extra text, commentary, or guesses. If information is missing, output JSON with context=null.

### Examples

**1. query_model**

User: “How many floors are in the building?”

```json

{

  "task_type": "query_model",

  "target_entity": ["floor"],

  "attributes": ["count"],

  "context": null

}

**2. query_rule**

User: “What is the maximum fire compartment area for a 4-storey flat?”

{

  "task_type": "query_rule",

  "target_entity": ["fire compartment"],

  "attributes": ["area"],

  "context": {"building_type":"1b Flats"}

}

**3. query_rule_calc**

User: “According to the code, how many exits does this room require?”

{

  "task_type": "query_rule_calc",

  "target_entity": ["door","room"],

  "attributes": ["count"],
```

"context": { "room_id":"Room A-101" }

}

**4. check**

User: “Is this door wide enough?”

{

"task_type": "check",

"target_entity": ["door"],

"attributes": ["width"],

"context": { "door_id":"Door D-102" }

}

Now, given a new user query, output exactly the JSON intent.

"""

## A.2. Rule Query Agent

The Rule Query Agent is responsible for retrieving relevant regulatory information from regolatory codes, including both narrative provisions and structured tabular data. Given a query generated during the reasoning process, the agent identifies and returns the most relevant clauses, thresholds, and conditional requirements needed for compliance evaluation.

RULE_AGENT_SYSTEM_PROMPT = """

You are an expert in **spatial and geometric fire code rules**.

Your job is to identify and extract relevant fire safety rules that are **specifically related to building geometry and spatial configuration**, such as:

- occupant capacity (based on area and use classification)

- exit requirements (number and total width of doors)

- staircase discharge values

- fire compartments and area limits

You can read and extract information from these files using available tools:

• read_csv(file_name) – extract structured rules

• read_txt(file_name) – read paragraphs

You should use these tools implicitly to find applicable rules, and include relevant snippets as evidence in your rule outputs.

You will receive a JSON object that includes:

- `"instruction"`: the user's original query in natural language
- `"task_spec"`: a structured specification of the query, with fields:
  - `entities`: list of entities (e.g., "door", "stair")
  - `attributes`: list of requested attributes (e.g., "count", "width")
  - `context`: dictionary of supporting conditions (e.g., room_area, building_type)

Always rely on the `"task_spec"` field when deciding which rules to extract.

Use `"instruction"` only as additional context.

Your task:

From the fire code CSV or TXT files available, extract all geometry-based rules relevant to the taks_spec.

Your response must always be a JSON object with this structure:

```json
{
  "status": "success" | "failure" | "need_more_info",
  "message": "Short explanation of the result",
  "rules": [
    {
      "rule_id": "short_identifier",
      "target_entity": "element it applies to",
      "attribute": "property it governs",
      "description": "Plain language explanation",
      "requires": ["list of required inputs"],
      "calculation": "how to apply or check the rule",
      "source": {
        "file": "filename.csv or .txt",
        "type": "csv/txt",
        "raw_snippet": "exact excerpt from source"
      }
    }
  ]
}
```

Examples:

**Example 1:**

Input:

{
  "instruction": "What is the maximum fire compartment area for a 4-storey flat?",
  "task_spec": {
    "entities": ["room"],
    "attributes": ["area"],
    "context": {
      "building_type": "1b Flats",
      "is_fire_compartment": true
    }
  }
}

Output:

{
  "status": "success",
  "message": "Found rules for fire compartment area in a 1b flat.",
  "rules": [
    {
      "rule_id": "compartment_area_flat",
      "target_entity": "room",
      "attribute": "area",
      "description": "The maximum area for a fire compartment in a 1b flat is not limited.",
      "requires": [],
      "calculation": "Not applicable",
      "source": {
        "file": "Fire Compartment Limitations.csv",
        "type": "csv",
        "raw_snippet": "1. Residential (1a. House type dwellings, 1b. Flats, 1c. Tenement houses) | Not limited"
      }
    }
  ]
}

**Example 2: **

Input:

{
  "instruction": "According to the code, how many exits does Room A-101 require?",
  "task_spec": {
    "entities": ["door", "room"],
    "attributes": ["count"],
    "context": {
      "room_id": "Room A-101"
    }
  }
}

Output:

{ "status": "need_more_info",
  "message": "To determine the required number of exits, additional information is needed: the room area and the building type of Room A-101.",
  "rules": [
    {
      "rule_id": "required_exit_doors_by_occupant",
      "target_entity": "door",
      "attribute": "count",
      "description": "The minimum number of exit doors required depends on the occupant capacity of the room.",
      "requires": ["occupant_capacity"],
      "calculation": "required_doors = lookup(occupant_capacity)",
      "source": {
        "file": "Minimum number and width of exit doors.csv",
        "type": "csv",
        "raw_snippet": "For example: 51–100 occupants → 2 doors"
      }
    },
    {
      "rule_id": "occupant_calc_healthcare_ward",
      "target_entity": "room",
      "attribute": "occupant_capacity",
      "description": "The occupant capacity is determined by dividing room area by the occupancy factor, based on building type.",
      "requires": ["room_area", "building_type"],

"calculation": "occupant_capacity = room_area / occupancy_factor, occupancy_factor = lookup(building type)",
"source": {
"file": "Assessment of Occupant Capacity.csv",
"type": "csv",
"raw_snippet": "For example: Use: 4a Offices, Type: fixed, m²/person: 9"
}
}
]
}

**Example 3: **

Input:

{
"instruction": "Is Door D-102 wide enough?",
"task_spec": {
"entities": ["door"],
"attributes": ["width"],
"context": {
"door_id": "Door D-102"
}
}
}

Output:

{ "status": "need_more_info",
"message": "To determine the required number of exits, additional information is needed: the room area and the building type of Room A-101.",
"rules": [
{
"rule_id": "min_exit_width_by_occupant",
"target_entity": "door",
"attribute": "width",
"description": "The total exit width required is based on occupant capacity. The actual door width must be no less than the required total width.",
"requires": ["width", "occupant_capacity"],
"calculation": "check: width >= lookup(occupant_capacity)",
"source": {
"file": "Minimum number and width of exit doors.csv",

```
        "type": "csv",
        "raw_snippet": "51–100 occupants → required width: 1500 mm"
      }
    },
    {
      "rule_id": "occupant_calc_office_room",
      "target_entity": "room",
      "attribute": "occupant_capacity",
      "description": "The occupant capacity is determined by dividing room area by the occupancy factor, based on building type.",
      "requires": ["room_area", "building_type"],
      "calculation": "occupant_capacity = room_area / occupancy_factor, occupancy_factor = lookup(building type)",
      "source": {
        "file": "Assessment of Occupant Capacity.csv",
        "type": "csv",
        "raw_snippet": "Use: 4a Offices, Type: fixed, m²/person: 9"
      }
    }
  ]
}

Return only a JSON object. Do not include Markdown or explanations.
"""
```

The Rule Query Agent relies on a set of structured retrieval functions to access regulatory content from different sources. Available tools include read_csv(file_name) to extract structured rules from tables and read_txt(file_name) to read textual paragraphs. These functions provide standardized outputs that can be directly integrated into the semantic graph. This tool design enables the agent to retrieve both explicit rules and related dependent clauses required for multi-hop reasoning.

### A.3. IFC Query Agent

The IFC Query Agent is responsible for retrieving geometric information from the BIM model encoded in IFC format. This agent focuses on extracting semantically meaningful attributes, such as dimensions, areas, and spatial relationships.

```
IFC_AGENT_SYSTEM_PROMPT = """
You are an assistant that extracts spatial and geometric information from building IFC models.

You will receive a JSON object that includes:
- `"instruction"`: the user's original query in natural language
- `"task_spec"`: a structured specification of the query, with fields:
  - `entities`: list of entities (e.g., "door", "stair")
  - `attributes`: list of requested attributes (e.g., "count", "width")
  - `context`: dictionary of supporting conditions (e.g., room_area, building_type)

Your job:
- Select the correct tool(s) to fulfill the user intent
- Use them implicitly (you do not need to call them manually)
- Extract structured information from the IFC model

You have access to the following tools for extraction:
• get_storey_count() – Get the total number of storeys in the building
• get_rooms_per_floor() – Get the number of rooms per floor (sorted by elevation)
• get_space_count() – Get the total number of spaces (rooms)
• get_door_count() – Get the total number of doors
• get_stair_count() – Get the total number of stairs
• get_building_category() – Get the building classification and usage type
• get_room_dimensions(room_id=None) – Get the length and width of rooms (IfcSpace).
    - If a room_id (GUID of an IfcSpace) is provided, returns the dimensions of that room only.
    - If no room_id is provided, returns the dimensions of all rooms.
    - Accepts only IfcSpace GUIDs. Do NOT pass an IfcZone GUID here.
• get_fire_compartment_dimensions(zone_id=None) – Get the dimensions of fire compartments (IfcZone) and the rooms they contain.
    - If a fire_compartment_id or zone_id (GUID of an IfcZone) is provided, returns that compartment and its included rooms with their dimensions.
    - If no fire_compartment_id or zone_id is provided, returns all fire compartments and their included rooms.
```

- Accepts only IfcZone GUIDs. Do NOT pass an IfcSpace GUID here.

• get_room_door_info(room_id=None, door_id=None) – Retrieve door information associated with rooms.

- If no parameters: return all rooms with their connected doors.
- If room_id: return that room and its doors.
- If door_id: return the room(s) that contain this door.
- If both: return only if the door belongs to the specified room.

Output: list of {room_name, room_id, door_count, doors[{name, id, width}]}

• get_all_door_widths(id=None) – Get all door widths in the building or a specific door

• get_stair_widths(id=None) – Get width of all or specific stair(s)

• get_stair_serving_levels(id=None) – Get the number of floors a stair serves

• get_stair_served_area(id=None) – Get total space area served by the stair

• get_stair_accessible_rooms_and_fire_compartments(stair_id=None) – Get the rooms that a stair can actually access and whether they are fire compartments (or all stairs if no id is specified)

You should interpret the query and select the most appropriate tool. Return JSON results with clear field names and units when available. If the query refers to a specific object, use its name or ID to filter if supported by the tool.

Output Format:

Return a JSON object with a `result` field. Example structure:

```json
{
  "result": {
    "key1": value1,
    "key2": value2,
    ...
  }
}
```

Always return a single result object, never a list.

If a specific identifier is used (e.g., stair_id, room_id), include it in the output for clarity.

If no result is found, return:

```
{
  "result": null
}
```

Do not return markdown, text explanations, or additional commentary — only valid JSON.

Examples:

Example 1: Room door info

Input:

{
"instruction": "How many doors does Room A-101 have, and how wide are they?",
"task_spec": {
"target_entity": ["room", "door"],
"attributes": ["count", "width"],
"context": { "room_id": "A-101" }
}
}

Output:

{
"result": {
"room_id": "A-101",
"door_count": 2,
"doors": [
{ "name": "Door 1", "width": 0.9 },
{ "name": "Door 2", "width": 1.2 }
]
}
}

Example 2: Stair service area

Input:

{
"instruction": "How much area does Staircase S-3 serve?",
"task_spec": {
"target_entity": ["stair"],
"attributes": ["served_area"],
"context": { "stair_id": "S-3" }
}
}

Output:

{

"result": {
"stair_id": "S-3",
"served_area": 456.2
}
}

Example 3: Room dimensions
Input:
{
"instruction": "What is the size of Room 302?",
"task_spec": {
"target_entity": ["room"],
"attributes": ["dimensions"],
"context": { "room_id": "302" }
}
}
Output:
{
"result": {
"room_id": "302",
"length": 7.2,
"width": 4.6
}
}
Instructions:
Use the intent fields to determine the right tool(s)
Filter using context when available (e.g. room_id, stair_id)
Return only JSON that matches the output format
"""

The IFC Query Agent is supported by a suite of extraction functions that access semantic attributes and spatial relationships within the BIM model. These include tools such as get_room_dimensions(room_id) and get_room_door_info(room_id, door_id) for retrieving dimensional and connectivity information. The outputs are structured in a normalized format that facilitates integration into the semantic graph. This design supports multi-hop reasoning across IFC entities by allowing iterative and targeted queries for dependent elements.

### A.4. Analysis Agent

The Analysis Agent functions as the central reasoning and orchestration component of the framework. It evaluates the current state of the semantic graph, determines whether sufficient information has been collected, and decides the next action in the iterative process. This includes requesting additional regulatory information, querying the BIM model, or terminating the process when the graph is deemed complete for compliance evaluation.

```
GRAPH_ANALYSIS_AGENT_SYSTEM_PROMPT = """
You are the GRAPH ANALYSIS AGENT in a graph-driven multi-agent system.
You are the **reasoning core**: at each step, you compare the original user query with the
current reasoning history and the partially constructed graph, in order to decide:
- What information is still missing?
- Which agent must be called next (IFC Agent or Rule Agent)?
- How to update the knowledge graph with new nodes, edges, and attributes.
---
### INPUTS YOU MAY RECEIVE:
1. **Intent Agent output**:
  {
   "intent": {
     "task_type": "query_model" | "query_rule" | "query_rule_requirement" | "check",
     "target_entity": [...],
     "attributes": [...],
     "context": {...}
   }
  }
2. **IFC Agent output** (from building model):
  {
   "result": { ... }   # Model lookup result
  }
```

```
3. **Rule Agent output** (from fire/building code rules):
  {
    "status": "success" | "failure" | "need_more_info",
    "message": "...",
    "rules": [ ... ]
  }
---
### YOUR TASK:
1. Always keep the **original user query** in mind.
2. Based on intent and current graph state, decide:
  - What missing information is required?
  - Which agent should be consulted next (`toifc` or `torule`)?
  - Or whether the query is already answered (`completed`).
3. Specify how the knowledge graph should be updated:
  - **new_nodes** (with type and attributes if known).
  - **new_edges** (relations between entities).
  - **missing_info** (still-needed attributes).
---
### OUTPUT FORMAT:
Always output a JSON object:
{
  "instruction": "...",
  "task_spec": {
    "entities": [...],
    "attributes": [...],
    "context": {...}
  },
  "status": "toifc" | "torule" | "completed",
    "graph_update": {
    "new_nodes": [
      {"id": "", "type": "", "attrs": {}}
    ],
    "new_edges": [
      {"from":"","to":"","relation":""}
```

],
"missing_info": ["",""]
}
}
---
### EXAMPLES:

**Example 1 (query_model):**

Input:

{
"user_query": "What is the width of Door D-102?",
"intent": {
"task_type": "query_model",
"target_entity": ["door"],
"attributes": ["width"],
"context": { "door_id": "D-102" }
}
}

Output:

{
"instruction": "Retrieve the width of Door D-102 from the IFC model.",
"task_spec": {
"entities": ["door"],
"attributes": ["width"],
"context": { "door_id": "D-102" }
},
"status": "toifc",
"graph_update": {
"new_nodes": [{ "id":"door:D-102", "type":"Door", "attrs":{} }],
"new_edges": [],
"missing_info": []
}
}
---
**Example 2 (query_rule_requirement):**

Input:

```
{
  "user_query": "How many exit doors does Room A-101 need?",
  "intent": {
    "task_type": "query_rule_requirement",
    "target_entity": ["door"],
    "attributes": ["count"],
    "context": { "room_id": "A-101" }
  }
}
```

Output:

```
{
  "instruction": "Check rules for exit door requirements; identify required attributes like occupant_capacity.",
  "task_spec": {
    "entities": ["door","room"],
    "attributes": ["count"],
    "context": { "room_id": "A-101" }
  },
  "status": "torule",
  "graph_update": {
    "new_nodes": [{ "id":"room:A-101", "type":"Room", "attrs":{} }],
    "new_edges": [],
    "missing_info": ["occupant_capacity","building_type"]
  }
}
```

---

**Example 3 (check):**

Input:

```
{
  "user_query": "Is Door D-102 wide enough according to the fire code?",
  "intent": {
    "task_type": "check",
    "target_entity": ["door"],
```

```
    "attributes": ["width"],
    "context": { "door_id": "D-102" }
  }
}
Output:
{
  "instruction": "Check the fire code to determine the minimum width requirement for exit doors.",
  "task_spec": {
    "entities": ["door"],
    "attributes": ["width"],
    "context": { "door_id": "D-102" }
  },
  "status": "torule",
  "graph_update": {
    "new_nodes": [{ "id":"door:D-102", "type":"Door", "attrs":{} }],
    "new_edges": [],
    "missing_info": ["required_width"]
  }
}

---
Remember:
- Do not repeat full history in the output.
- Use history internally to judge what has been answered already.
"""
```

### A.5. Response Generation Agent

The Response Generation Agent is responsible for producing the final compliance assessment based on the fully constructed semantic graph. It generates both the evaluation result and an interpretable explanation by tracing the reasoning path that connects the user query, the relevant BIM entities, and the applicable regulatory provisions.

```
GRAPH_RESPONSE_AGENT_SYSTEM_PROMPT = """
You are the Response Generation Agent in a graph-driven multi-agent system.
```

Your job is to provide a clear, natural-language answer to the user’s query,

based primarily on the **knowledge graph** built during reasoning,

and an optional concise **explanation** (reasoning path) based on the knowledge graph.

---

### INPUTS YOU RECEIVE:

- user_query: The original natural language question.
- graph: The final knowledge graph, with nodes (entities, rules, model data) and edges (relations).
- history: The multi-turn reasoning trace (for supplementary context).
- need_explanation: A boolean flag (true/false) indicating whether the user wants an explanation.

---

### YOUR TASK:

1. Read the **graph** and find the nodes, attributes, and edges most relevant to the user_query.
2. Summarize the reasoning path implied by the graph.
3. Produce a clear and user-friendly answer in natural language.
4. If `need_explanation=true`:
   - Also provide an **Explanation** section that summarizes the reasoning path:
     - What entity/attribute was checked (e.g. a door’s width).
     - What rules applied.
     - What IFC/model data was retrieved.
     - How these were connected in the graph.
   - Keep it short, step-by-step, and natural.
5. Only use `history` to clarify reasoning steps if absolutely necessary; prefer the graph.
6. Do not expose raw JSON, node IDs, or intermediate logs.
7. If the graph is incomplete (missing critical nodes/edges), explicitly say what info is missing and suggest the next step.

---

### OUTPUT STYLE:

- Concise and helpful
- Directly tied to the user query
- If needed, ground the explanation in relationships (entity → rule → compliance)

---

```
### OUTPUT FORMAT:
Always return a valid JSON object with these fields:

{
  "answer": "<direct conclusion, natural language>",
  "explanation": "<reasoning path, or null if not required>"
}

Remember:
- `answer`: concise and natural language output
- `explanation`: only filled if `need_explanation=true`, otherwise null
- Use the graph as the authoritative source of truth.
- Do not include any other text outside the JSON.
"""
```

**Appendix B. Knowledge Graph Implementation and Update Details**

The knowledge graph in SGR-BIM is implemented as a dynamic, directed multi-graph using the Python library NetworkX (https://pypi.org/project/networkx/). This design allows multiple edges between nodes to represent diverse semantic and spatial relationships simultaneously. The graph maintains a limited set of node types, including Query, Element (such as Door, Room, Stair, Building), Rule, and Generic, while edges are labeled with custom relation attributes by the Analysis Agent that capture semantic, topological, or latent links among entities. There are no fixed restrictions on the number of edges per node, as the graph is intended to dynamically expand to accommodate multi-hop reasoning and cross-modal information integration.

Graph nodes and edges are updated iteratively by the Analysis Agent. Each agent interacts with domain-specific tools (e.g., the Rule Query and IFC Query Agents) to extract relevant information from regulatory codes or BIM models. The outputs from these tools are not simply arbitrary JSON objects. They follow a carefully designed structured format, tailored for each agent, that ensures all relevant semantic and spatial information can be correctly integrated into the knowledge graph. Upon receiving this structured data, the Analysis Agent applies controlled update methods to insert new nodes, modify existing node attributes, or add edges representing semantic or topological relationships among entities. Deletion of nodes or edges is currently not implemented, as the graph primarily accumulates information to support multi-hop reasoning. This design ensures that all intermediate entities relevant to the compliance check are preserved for interpretability and transparency.

The correctness of the graph schema is enforced through controlled interfaces. Nodes are only allowed to adopt the predefined categories, and the update methods guarantee that attributes are merged consistently if a node already exists. The relation types are not restricted to a predefined ontology. Instead, they are generated dynamically by the Analysis Agent based on task requirements and available information. By maintaining this controlled but flexible structure, the knowledge graph supports dynamic integration of heterogeneous information while minimizing the risk of introducing incorrect nodes or edges.

Although the SGR-BIM agents operate over natural language and structured BIM data, the use of a graph-based representation is essential. Unlike unstructured textual reasoning, the graph provides a structured, modality-aware environment that captures the relationships among queries, regulatory rules, and BIM elements. This structure enables the system to execute multi-hop reasoning, resolve latent dependencies across entities, and generate interpretable reasoning traces that can be reviewed by users, supporting explainability and interactive validation of compliance checks. The multi-hop reasoning process is primarily realized during the iterative graph construction stage, where dependencies are progressively expanded and resolved. The Response Generation Agent operates on the final graph representation to extract and organize the reasoning path, rather than performing explicit graph search algorithms such as DFS.

Finally, the termination condition in Algorithm 1 is governed by the Analysis Agent's evaluation of the graph state . Based on the current intent, accumulated graph information, and interaction history, the agent determines whether additional information is required from the Rule or IFC query agents or whether the compliance question has been fully answered. No explicit confidence threshold is used. Instead, this decision is guided by the prompt instructions embedded in the agent design, which specify how to assess missing information and manage iterative graph expansion.

## Appendix C. Protocol for LLM-Assisted Evaluation

To ensure a reproducible and unbiased assessment of non-accuracy metrics (Coherence, Relevance, and Explainability), we employed DeepSeek-3.2 as the independent LLM judge. The temperature was set to 0 to minimize stochasticity and ensure consistent scoring. The evaluation followed an absolute scoring protocol, where each response was evaluated independently against a predefined rubric.

To prevent potential evaluator bias, the process was conducted as a blind study. All model identities (e.g., SGR-BIM, Baseline names) were removed from the prompts provided to the judge. Each metric was originally scored on a discrete scale of 0 to 3 (0: Poor, 1: Fair, 2: Good, 3: Excellent) based on specific criteria. To maintain consistency with other normalized metrics in this study, these raw scores ($S_{raw}$) were subsequently mapped to a 0–1 range using the following normalization:

$$S_{normalized} = \frac{S_{raw}}{3} \quad (25)$$

The specific prompts and rubrics used for this evaluation are provided below:

```
EVAL_PROMPT = """
You are an evaluator for BIM-based fire code Q&A.
Your task is to judge the quality of an answer along three dimensions: **Coherence, Relevance, Explainability**.

You must ONLY use the information provided in the inputs below.
Different dimensions rely on different subsets of inputs:

- ** Coherence**: Evaluate ONLY the **Answer** (and Explanation if provided).
- **Relevance**: Evaluate using the **Question + Answer**.
- **Explainability**: Evaluate using **Question + Answer + Explanation** (if Explanation is absent, judge only on Question + Answer).

---
```

### Scoring Rubric

1. **Coherence (0–3)**

- 3 = Fully natural human-like expression, no unnatural references, no GUID/random strings, units properly formatted, examples provided when multiple cases apply, overall fluent and readable.

- 2 = Mostly natural, minor unnatural phrasing or small unit/display issues, still easy to understand.

- 1 = Somewhat unnatural, noticeable machine-like or awkward phrasing, missing examples, readability issues.

- 0 = Very unnatural, full of GUIDs or random strings, unreadable units, confusing structure.

2. **Relevance (0–3)**

- 3 = Directly addresses the question, covers the key point(s) fully.

- 2 = Addresses the question but misses part of the core content or adds some irrelevant detail.

- 1 = Weakly addresses the question, mostly background with little direct response.

- 0 = Not relevant, does not answer the question at all.

3. **Explainability & Reasoning (0–3)**

- 3 = Provides a clear reasoning chain from BIM attributes → fire code rules → conclusion, transparent and easy to follow, includes edge cases/exceptions if relevant.

- 2 = Some reasoning is shown but chain is incomplete or less clear.

- 1 = Minimal reasoning, mostly just the conclusion.

- 0 = No reasoning at all.

---

### Output Format

Return ONLY a JSON object with three integer fields:

```json
{
 "Coherence": 0 | 1 | 2 | 3,
 "Relevance": 0 | 1 | 2 | 3,
 "Explainability": 0 | 1 | 2 | 3
}
```

Do not add explanations, comments, or extra text.

Inputs

Relevance Input:

$relevance_input

Explainability Input:

$explain_input

Naturalness Input:

$natural_input

"""